\documentclass[journal]{IEEEtran}
\ifCLASSINFOpdf
\else
\fi

\usepackage{graphicx}

\usepackage{amssymb}

\usepackage{amsthm}

\usepackage{amsmath}

\usepackage{lineno}

\usepackage{caption} 

\usepackage{array} 

\usepackage{booktabs} 

\usepackage{bm}

\usepackage{setspace} 
\usepackage{footnote}

\usepackage{inputenc}
\usepackage{algorithm}  
\usepackage{algpseudocode}  
\usepackage{amsmath}  

\usepackage[table]{xcolor}
\usepackage{mathrsfs}

\usepackage{color}
\usepackage[colorlinks,linkcolor=red,anchorcolor=red,citecolor=green]{hyperref}

\begin{document}
%
\title{X-LineNet: Detecting Aircraft in Remote Sensing Images  by a pair of Intersecting Line Segments}
%
%
%


\author{Haoran~Wei,~\IEEEmembership{}
        Yue~Zhang*,~\IEEEmembership{}
        Bing~Wang,~\IEEEmembership{}
        Yang~Yang,~\IEEEmembership{}
        Hao~Li
        and~Hongqi~Wang~\IEEEmembership{}

\thanks{Haoran~Wei and Bing~Wang are with the Aerospace Information Research Institute, Chinese Academy of Sciences, Beijing 100190, China, the School of Electronic, Electrical and Communication Engineering, University of Chinese Academy of Sciences, Beijing 100190, China and the Key Laboratory of Network Information System Technology (NIST), Aerospace Information Research Institute, Chinese Academy of Sciences, Beijing 100190, China e-mail: weihaoran18@mails.ucas.ac.cn, wangbing181@mails.ucas.ac.cn.}
\thanks{Yue~Zhang, Hao~Li and Hongqi~Wang are with the Aerospace Information Research Institute, Chinese Academy of Sciences, Beijing 100190, China and the Key Laboratory of Network Information System Technology (NIST), Aerospace Information Research Institute, Chinese Academy of Sciences, Beijing 100190, China e-mail: zhangyue@aircas.ac.cn.}

\thanks{Yang~Yang is with the Unit 31008 of the PLA.}

}

\maketitle

\begin{abstract}
	
Motivated by the development of deep convolution neural networks (\textit{DCNNs}), tremendous progress has been gained in the field of aircraft detection. These \textit{DCNNs} based detectors mainly belong to top-down approaches, which first enumerate massive potential locations of objects with the form of rectangular regions, and then identify whether they are objects or not. Compared with these top-down approaches, this paper shows that aircraft detection via bottom-up approach still performs competitively in the era of deep learning. We present a novel one-stage and anchor-free aircraft detection model in a bottom-up manner, which formulates the task as detection of two intersecting line segments inside each target and grouping of them without any rectangular region classification. This model is named as \textit{X-LineNet}. With simple post-processing, \textit{X-LineNet} can simultaneously provide multiple representation forms of the detection result: the horizontal bounding box, the rotating bounding box, and the pentagonal mask. The pentagonal mask is a more accurate representation form which has less redundancy and can better represent aircraft than that of rectangular box. Experiments show that \textit{X-LineNet} outperforms state-of-the-art one-stage object detectors and is competitive compared with advanced two-stage detectors on both \textit{UCAS-AOD} and \textit{NWPU VHR-10} open dataset in the field of aircraft detection.	

\end{abstract}

\begin{IEEEkeywords}
Aircraft detection, Bottom-up methods, Deep convolution neural network, Remote sensing image.
\end{IEEEkeywords}

%
\IEEEpeerreviewmaketitle

\begin{figure*}[t]
	\centering
	\begin{minipage}{9cm}
		\includegraphics[width=9cm]{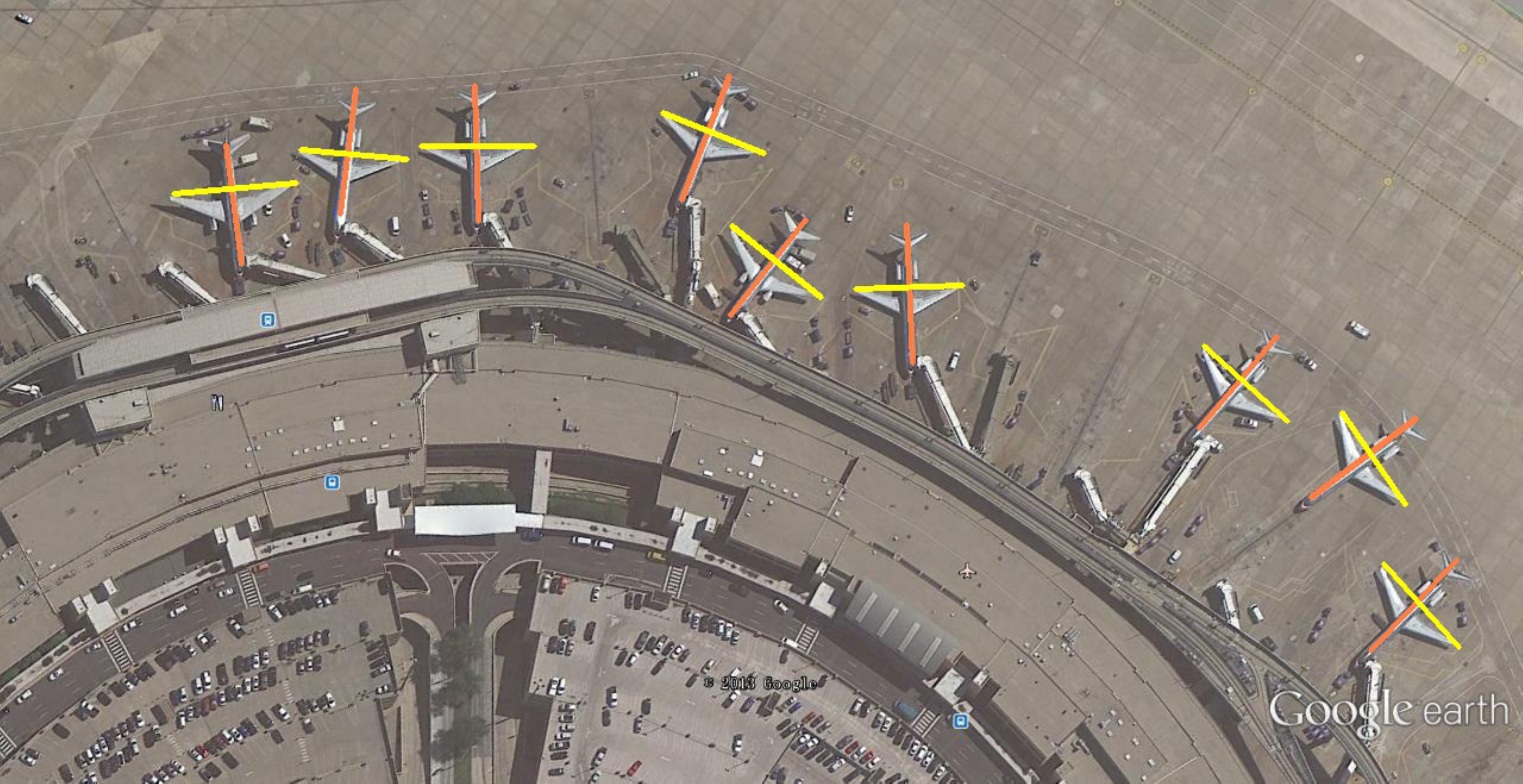}
		
	\end{minipage}
	\begin{minipage}{9cm}
		\includegraphics[width=9cm]{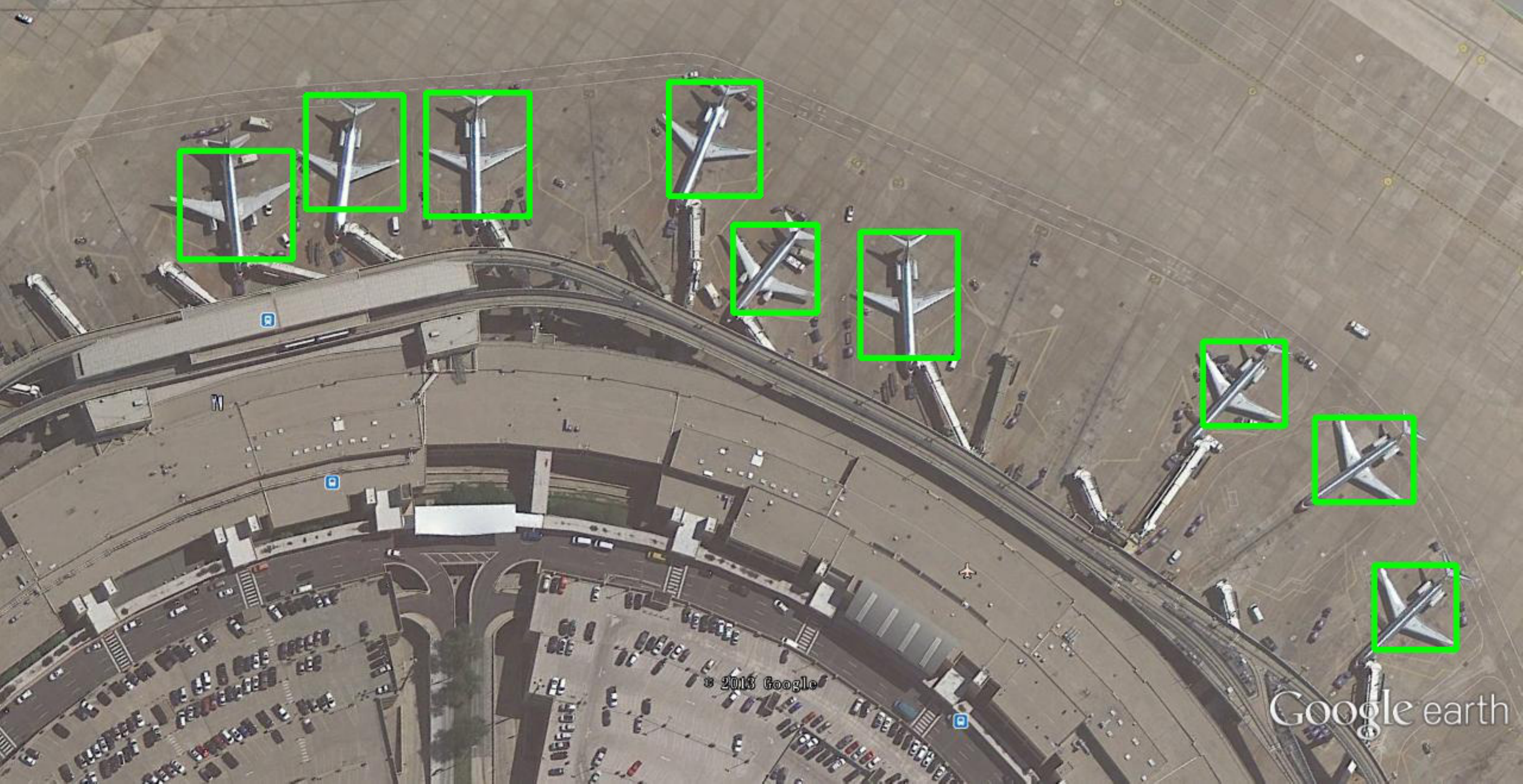}
		
	\end{minipage}

	\vspace{1 ex}
	\begin{minipage}{9cm}
		\includegraphics[width=9cm]{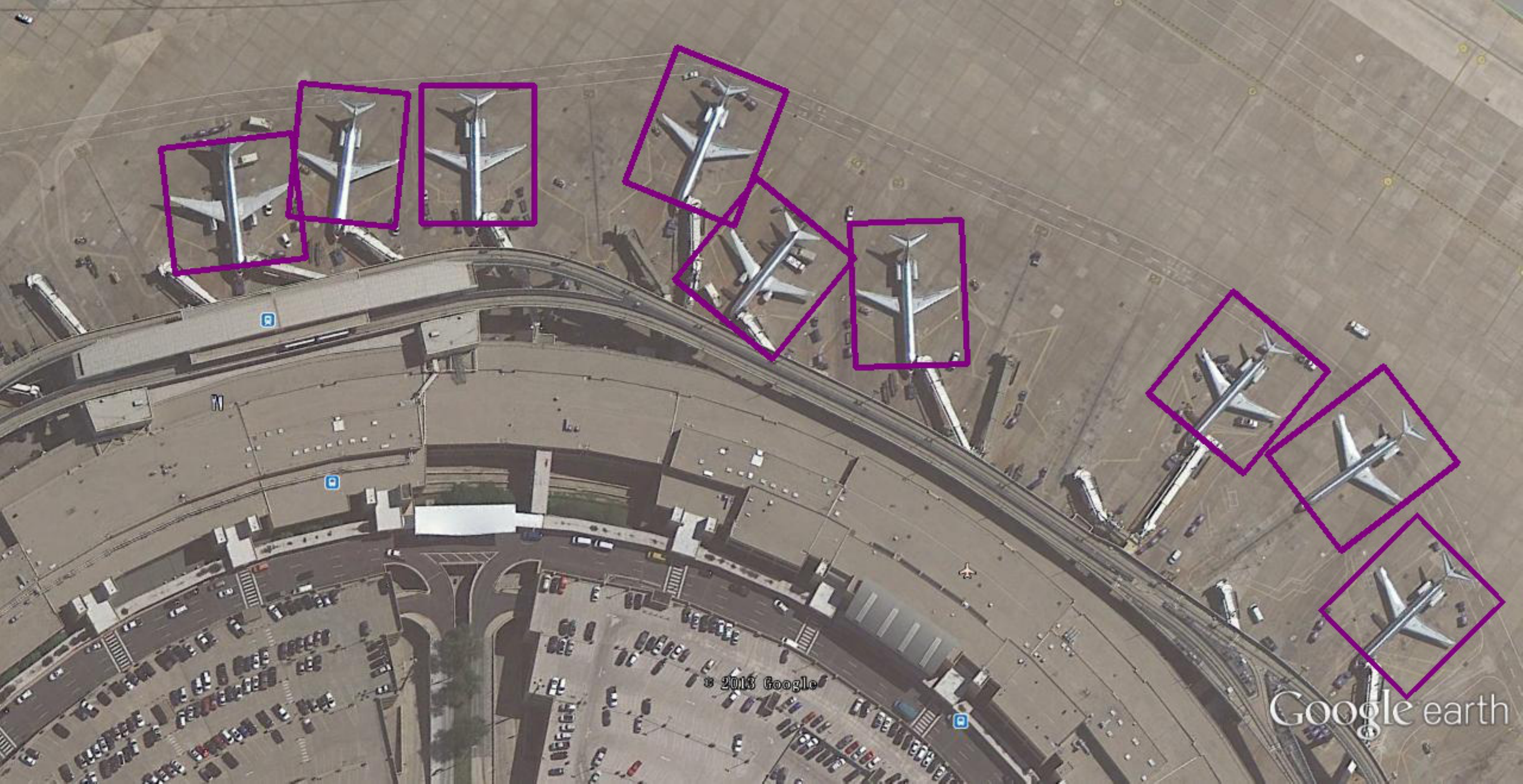}
	\end{minipage}
	\begin{minipage}{9cm}
		\includegraphics[width=9cm]{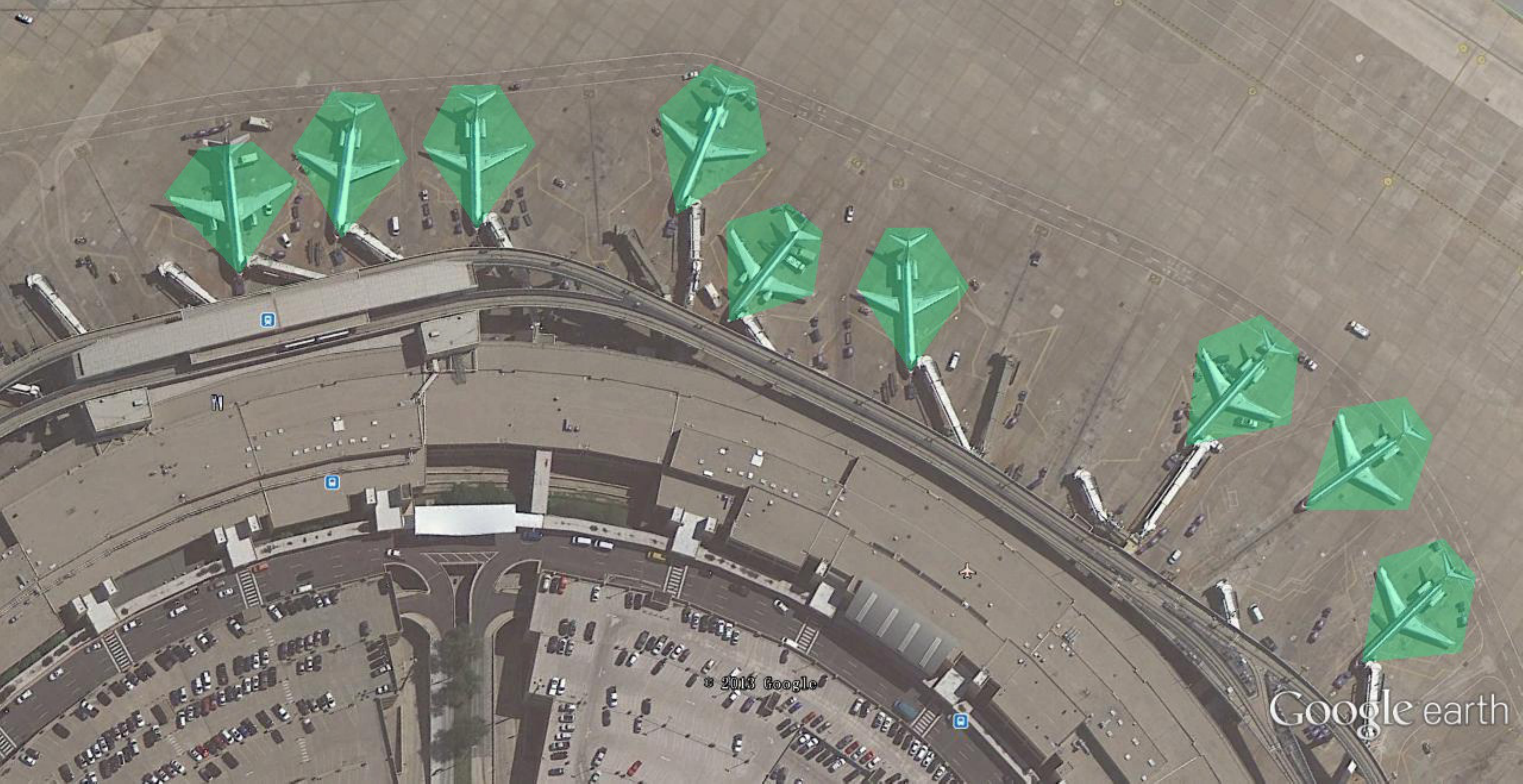}
	\end{minipage}
	\caption{We predict a pair of line segments to obtain the aircraft's bounding boxes. The left upper figure is the line segments inside the aircraft predicted by our model , the right upper figure is the horizontal bounding boxes, the left bottom one is the rotating bounding boxes, and the last one is the pentagonal mask.}
	\label{Figure 1}
\end{figure*}

\section{Introduction}
%
%
%
%
\IEEEPARstart{D}{riven} by the rapid development of remote sensing technologies, remote sensing images with finer resolution and clearer texture can be easily accessed by the modern airborne and space borne sensors. In response to the demand for automatic analysis of remote sensing data, object detection has been widely researched, among which aircraft detection occupies an important part owning to the successful application in the field of military and air transportation. However, aircraft detection remains a great challenge on account of the need for accurate and robust detection. Moreover, the complex background and noise, as well as the variation of spatial resolution of remote sensing images, increase the difficulty of this task.

Conventional methods in aircraft detection mainly depend on manually engineered features. Cheng et al.\cite{cheng2013object} develop a discriminatively trained model for extracting Histogram of Oriented Gradient (\textsl{HOG})\cite{dalal2005histograms} feature pyramids from multiple scale. Then threshold operation is performed on the response of the model to judge the presence of aircraft. Bai et al.\cite{bai2014vhr} select structural feature description as input of ranking Support Vector Machine(\textsl{SVM})\cite{cortes1995support} to identify the existence of object. Besides, Rastegar et al.\cite{rastegar2009airplane} build an aircraft detection system in combine \textsl{Wavelet} features with \textsl{SVM} classifiers. Though these conventional methods are effective in specific scenes, the generalization capability of them needs to be further improved in practice.

Inspired from the success of deep convolution neural networks (\textit{DCNNs})\cite{lecun1998gradient}\cite{krizhevsky2012imagenet}, various methods based on \textit{DCNNs} have been proposed in the research field of aircraft detection\cite{ding2018light}\cite{zhang2016weakly}\cite{cao2016region}. Most of these algorithms follow the principle of top-down detectors\cite{ren2015faster}\cite{liu2016ssd}\cite{lin2017focal}. Because they can automatically extract features through the backbone networks\cite{simonyan2014very}\cite{he2016deep}, the accuracy as well as robustness is greatly improved compared with manually engineered features.  Cao et al.\cite{cao2016region} use selective search methods to generate a large set of region proposals, and then send them into classifiers as well as modify the final bounding boxes in a regression manner. Similarly, Ding et al.\cite{ding2018light} adopt a region proposal network\cite{ren2015faster} module to generate the proposals from a large set of anchor boxes\cite{ren2015faster}, which are candidate boxes with fixed size densely distributed over the feature map. 

Although satisfactory results have been achieved by these top-down approaches, there are still some limits of them. First, this type of method converts aircraft detection into task of classifying rectangular regions, but the rectangular region is not an accurate representation for aircraft: rectangular box not only  includes many redundant pixels which  belong to background actually, but also will ignore some  details, such as  direction contexts of aircraft. Besides, top-down aircraft detectors rely on anchor mechanism heavily. In order to locate the targets with no omission, a large set of anchor boxes with different size and ratio are densely placed over the feature maps. While only tiny set of boxes will have an overlap with ground truth, leading to the imbalance in positive and negative samples, which hinders the convergence of the network during training. In addition, the targets of regression in anchor-based detectors are the candidate boxes with the overlap with ground truth above $0.7$ typically, which doesn't truly understand the concrete aircraft's visual grammar information\cite{zhou2019bottom}\cite{felzenszwalb2009object}\cite{girshick2011object}. Consequently, numerous redundant bounding boxes will be generated, which require a series of post-processing such as Non-Maximum Suppression(\textit{NMS}) to filter the final bounding boxes. It is complex and wasteful. Moreover, many hyper parameters are introduced related to anchors, such as sizes and aspect ratios of anchor boxes. All of these require a lot of prior knowledge to deploy.

Considering the limits mentioned above in top-down anchor-based approaches,  we promote a bottom-up anchor-free model named \textit{X-LineNet} to detect aircraft in this paper. \textit{X-LineNet} is inspired by \textit{CornerNet}\cite{law2018cornernet} which detects objects via predicting and grouping paired corner points. However, \textit{CornerNet} is not suitable for the task of aircraft detection in remote sensing images: a single optical image may contain a large amount of aircraft, and this situation will frequently lead to errors of grouping points detected in \textit{CornerNet}. Unlike \textit{CornerNet}, our \textit{X-LineNet} formulates this task as the detection of two line segments inside each aircraft and clustering of linear segments. These two sets of target segments are defined as a pair of linear area from the head of the aircraft to the tail part as well as area between left and right wings. \textit{X-LineNet} first predicts two heatmaps corresponding to two set of line segments respectively, and then groups these line segments according to their geometric relationships. As the result of dimension ascending from point to line, the difficulty of points grouping in \textit{CornerNet} is solved. In addition, \textit{X-LineNet} abandons the anchor mechanism in top-down detectors\cite{ren2015faster}\cite{liu2016ssd}\cite{redmon2017yolo9000}\cite{redmon2018yolov3} and successfully solves the problems related to anchors aforementioned. Except for the commonly used horizontal bounding boxes and rotating bounding boxes, \textit{X-LineNet} can also provide a new output form of the detection result as pentagonal mask, which has richer semantic description of each aircraft target, such as the refined position and direction information. Fig. \ref{Figure 1} illustrates these results.

Our contributions and innovations are as follows:

(1).We promote a novel aircraft detection model named \textit{X-LineNet}, which goes in a bottom-up manner. Different from the dominant top-down methods in aircraft detection, this paper shows that bottom-up approach can still perform competitively in this filed during the era of deep learning.

(2).\textit{X-LineNet} transforms the task of aircraft detection in remote sensing images to predicting and grouping of paired intersecting line segments, which enables the network to learn richer and specific visual grammar information.

(3).Besides horizontal and rotating rectangular bounding boxes of detection result, one more accurate form as the pentagonal mask can be generated within a single network simultaneously. It contains less redundant context, and can provide richer semantic description of an aircraft, such as the direction and more accurate localization.

The rest of the paper is organized as follows: In  {\color{red}\textit{Section \ref{section:2}}}, we introduce the related works done by researchers before and basic principle in our method. The details of our network and algorithms are shown in {\color{red}\textit{Section \ref{section:3}}}. We place our experiments results and analysis in {\color{red}\textit{Section \ref{section:4}}}. At last, our work is summarized and concluded in {\color{red}\textit{Section \ref{section:5}}}.

\begin{figure*}[t]
	\centering
	\begin{minipage}{5.8cm}
		\includegraphics[width=5.8cm]{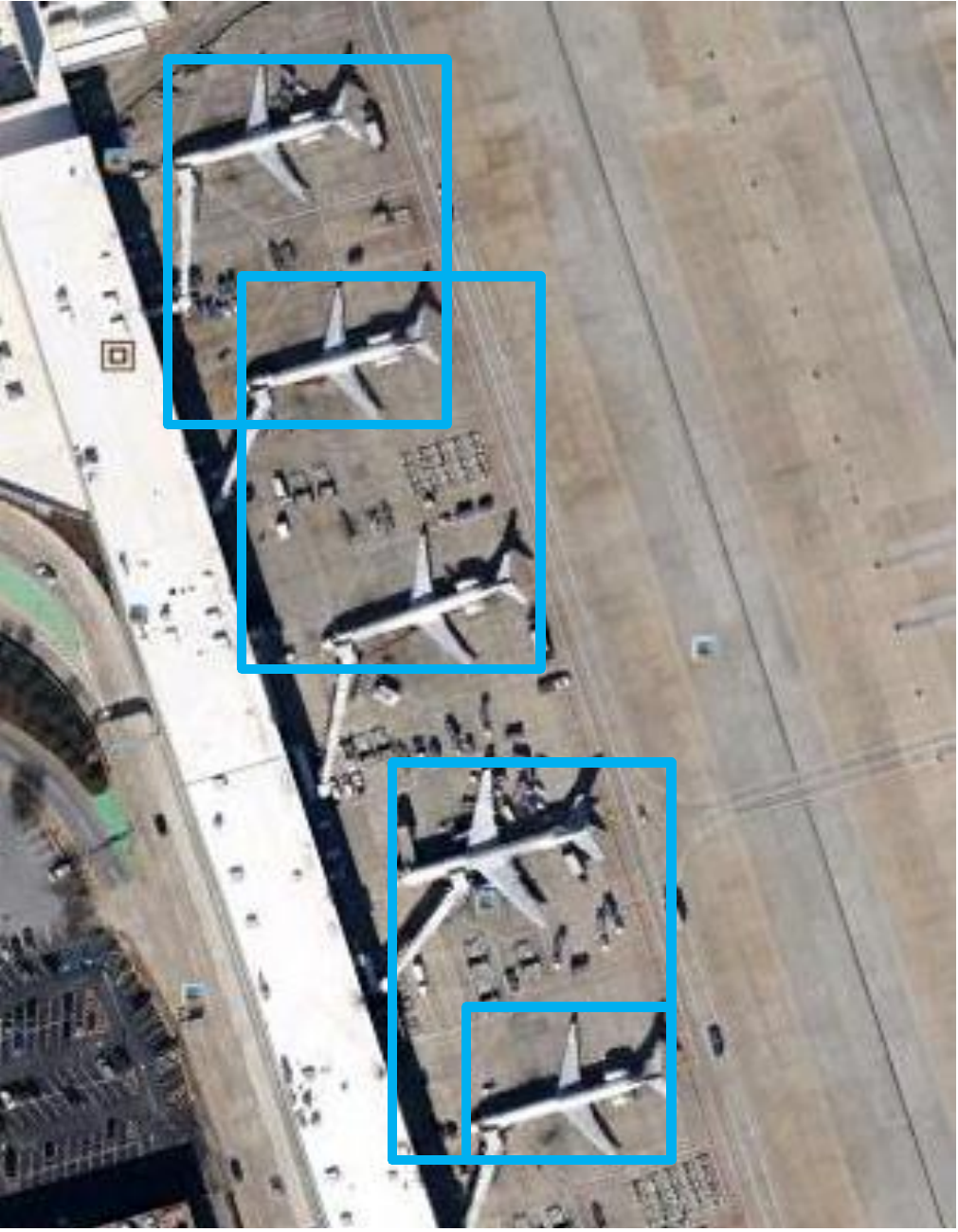}
		\caption*{(a)}
	\end{minipage}
	\begin{minipage}{5.8cm}
		\includegraphics[width=5.8cm]{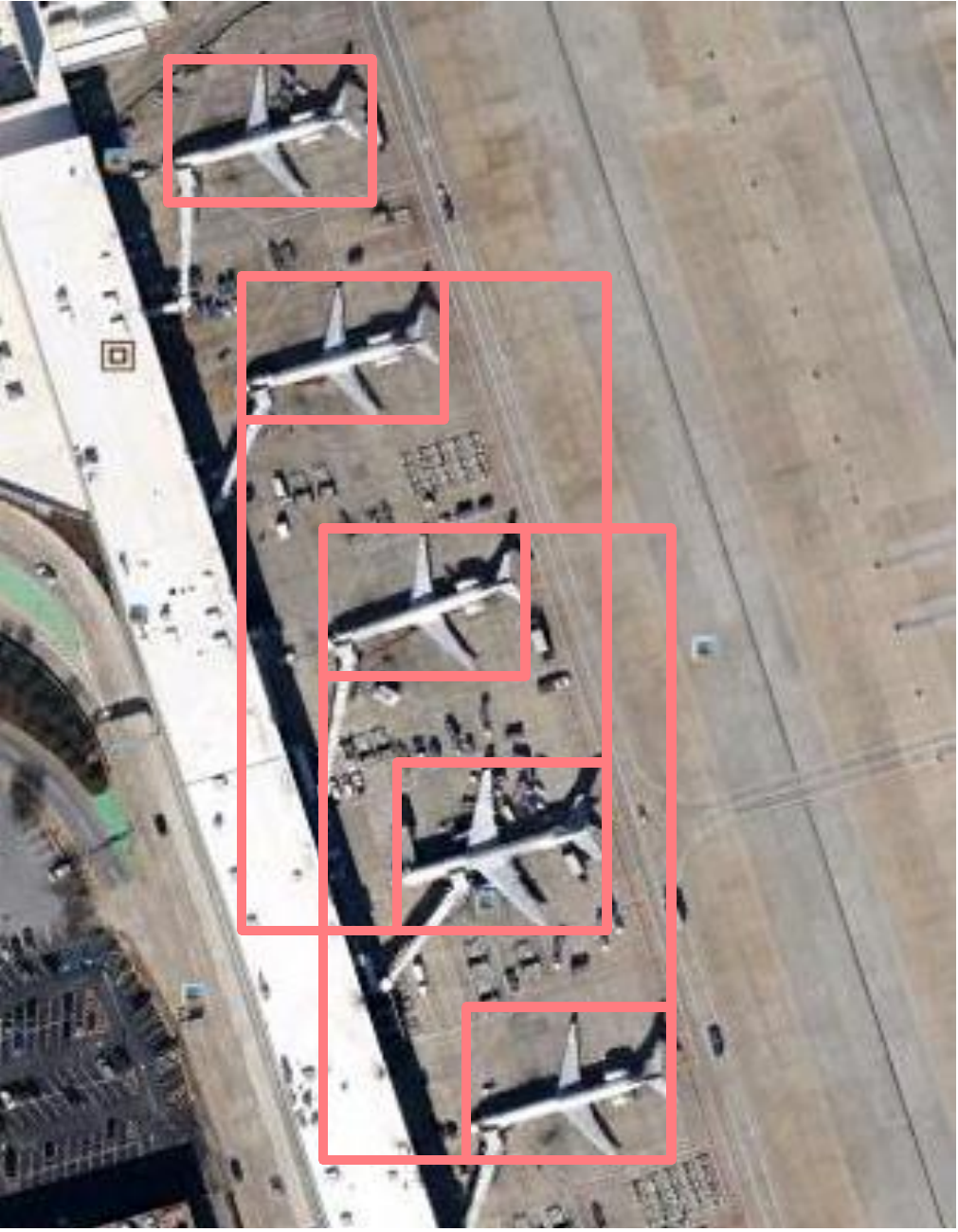}
		\caption*{(b)}
	\end{minipage}
	\begin{minipage}{5.8cm}
		\includegraphics[width=5.8cm]{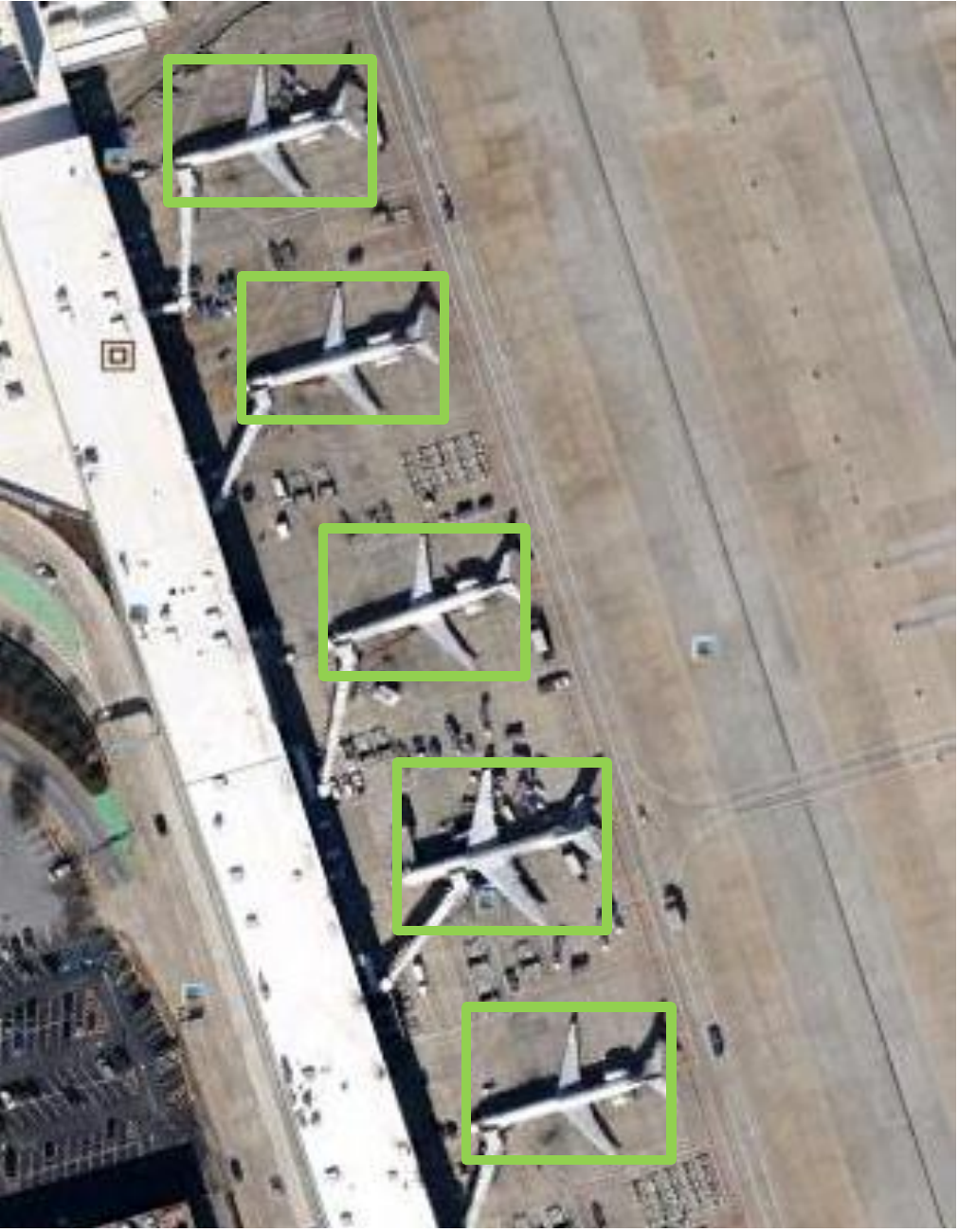}
		\caption*{(c)}
	\end{minipage}
	\caption{The figure (a) shows the \textit{CornerNet's} effect, figure (b) for \textit{ExtremeNet} and figure (c) for \textit{X-LineNet}.}
	\label{Figure 2}
\end{figure*}

\section{Related Works} \label{section:2}

\subsection{Top-down Aircarft Detectors with Anchors}
Top-down frameworks based on anchor mechanism have dominated the field of aircraft detection\cite{ding2018light}\cite{zhang2016weakly}\cite{cao2016region} recently. Generally, these methods can be classified into two-stage detectors and one-stage detectors.

\subsubsection{Two-stage Detectors}

Two-stage detectors decompose the task of detection into subtasks of generating region proposals and classifying of these candidate boxes. In the early \textit{R-CNN}\cite{girshick2014rich}, region proposals are produced under the algorithm of selective search, which can be quite inefficient as the generation of many redundant proposals. In order to lift the efficiency, \textit{SPP}\cite{he2015spatial} and \textit{Fast R-CNN}\cite{girshick2015fast} design a special pooling layer to improve it, but both of them are not end-to-end trainable. Later, a region proposal network(\textit{RPN})\cite{ren2015faster} is proposed in \textit{Faster R-CNN}\cite{ren2015faster}. Mechanism of \textit{RPN} can be illustrated in detail as below. 

After the feature maps extracted by the backbone\cite{simonyan2014very} from the original image, a large set of anchor boxes can be produced by densely placing rectangular boxes with different size and ratio. On accout of filtering these numerous anchors, \textit{NMS} and scoring mechanism within \textit{RPN}\cite{ren2015faster} are performed to generate the final regions of candidate. Final detection results can be accessed after the classification and modification of these proposals in parallel. Benefited from the adoption of \textit{RPN}\cite{ren2015faster}, the network can be trained end-to-end and the accuracy of detection results greatly increased.

\subsubsection{One-stage Detectors}

In pursuit of computation efficiency and inference speed, one-stage detectors get on detection within a single network away with region proposal module. Initially, \textit{YOLOv1}\cite{redmon2016you} without anchor mechanism can't provide accuracy comparable to that of two-stage detectors. Later, anchor methods are also extensively utilized in one-stage detectors\cite{redmon2017yolo9000}\cite{liu2016ssd}\cite{lin2017focal} to improve the accuracy, massive anchors with different scales are densely placed within certain layers of the network. 

Besides the gain in accuracy, there are some problems accompanied by the bringing in anchor mechanism in one-stage detectors. First, a large set of anchors from multi scale can be produced to supply sufficient location information, while only a tiny set of them are positive samples with the overlap with ground truth above $0.7$ or $0.5$ typically. This phenomenon leads to the imbalance in positive and negative samples, which may prolong the process of training. Besides, many hyper parameters are introduced, for example, concrete settings for size and radio of anchor boxes in certain layers, which bring in much inconvenience to adjust them carefully.

On account of  drawbacks of anchor mechanism mentioned above, there has been many works on modifying it. In \textit{RetinaNet}\cite{lin2017focal}, \textit{Focal Loss}\cite{lin2017focal} is promoted to avoid the issue of one-stage detector's overwhelming by negative samples: In the process of generating anchors, weights are adjusted according to the distributions of samples. Furthermore, \textit{MetaAnchor}\cite{yang2018metaanchor} is proposed to decrease the reliance of anchors' settings on prior knowledge: Benefited from the sub network of anchor function generator, hyper parameters of anchor boxes can be learned dynamically and generalization of the model is greatly enhanced.

Our approach falls into category of one-stage detectors as the process of region proposal is discarded and accomplishes the task of detection away with anchor mechanism. It is worth noting that the target space of searching is reduced from the $O(w^2h^2)$ to $O(wh)$ by abandoning anchors.

\begin{figure*}[t]
	\centering
	\includegraphics[width=18.5cm]{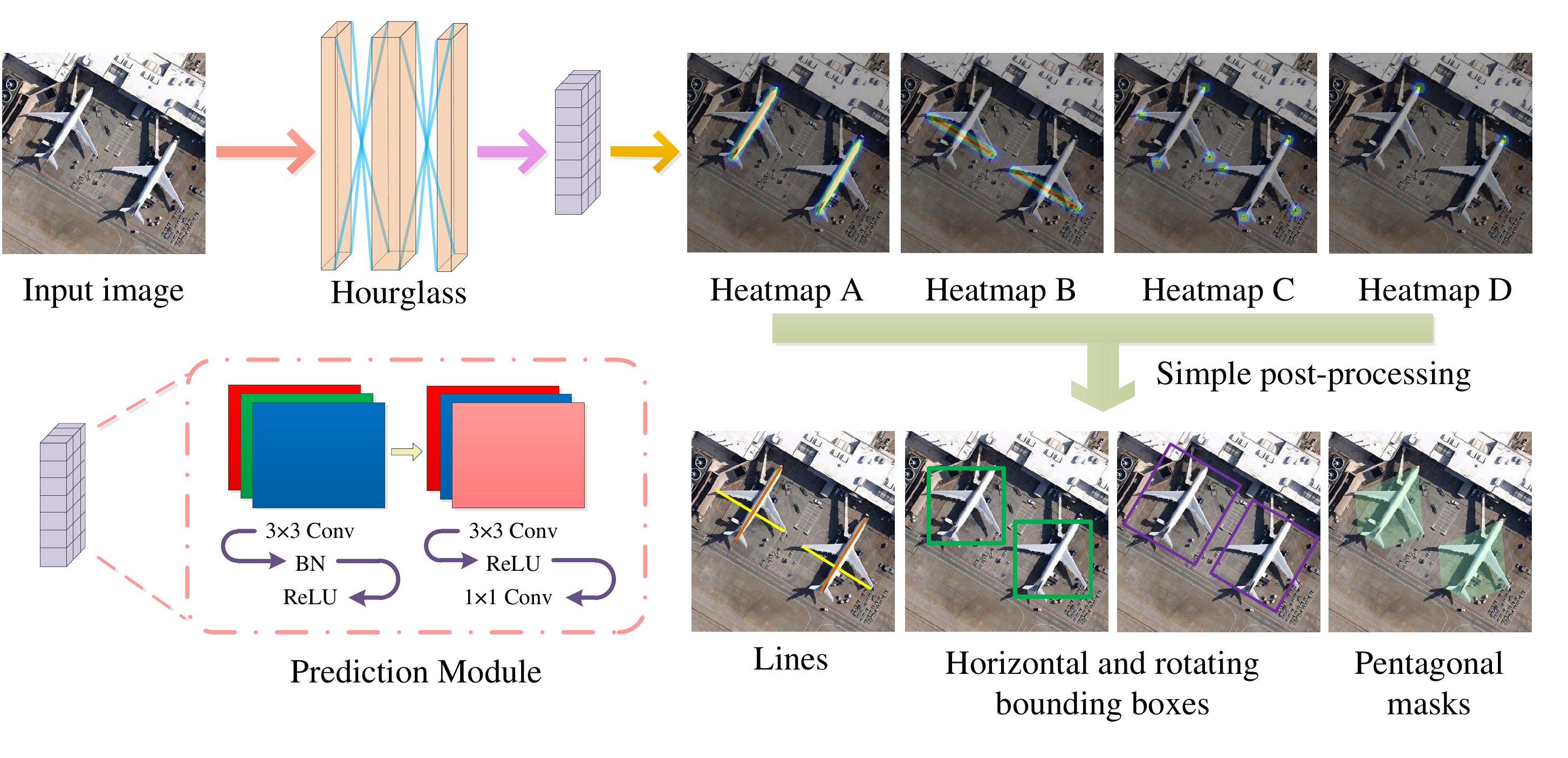}
	\caption{Illustration of our framework. The "Conv" is convolution layer, "BN" is Batch Normalization layer and "ReLU" is ReLU layer.}
	\label{Figure 3}
\end{figure*}

\subsection{Bottom-up Detectors Based on DCNNs}

In recent times, the research of detection in an \textit{Anchor-Free} and bottom-up manner becomes popular. \textit{CornerNet}\cite{law2018cornernet} and \textit{ExtremeNet}\cite{zhou2019bottom} are two representative detectors of this new approach. Their idea is to transform the object detection task into the processes of predicting and grouping keypoints, which can be seen as detecting objects in a bottom-up way, and state-of-the-art results has been attained by them in \textit{MS COCO}\cite{lin2014microsoft} benchmark. While it is not suitable to apply them into aircraft detection in remote sensing images for the reasons followed:

\textit{CornerNet's}\cite{law2018cornernet} clustering algorithm draws on the method of Newell et al.\cite{newell2017associative}. More specifically, they group each corner points detected by embedding a vector on them. The basic assumption is that the distance (\textit{Euclidean Distance}) between two corner points of the same object is small, and large between different objects. However, this process is not applicable to an intensive occasion: with the large number of the same class objects densely distributed in a single image, such as remote sensing aircraft image, this algorithm has difficulty in converging and results in embedding errors frequently.

The grouping method of ExtremeNet\cite{zhou2019bottom} is to match 4 extreme points with the center point of objects, and basic rules are based on the assumption that if the geometric center point of 4 extreme points match the predicted center point then the same object they should  belong to. However, this method is still unable to be applied to detect aircraft due to the fact that in many cases airplanes may be spatially symmetric in remote sensing images.

Our model converts detection of points to line segments, which effectively solves the clustering problems that present \textit{Anchor-Free} and bottom-up detectors fall into in the task of aircraft detection. Fig. \ref{Figure 2} shows representative contrast results between \textit{X-LineNet} and other two bottom-up detectors aforementioned.

\section{X-LineNet}\label{section:3}
\subsection{Overview} \label{section: 3.1}

\textit{X-LineNet} detects each aircraft as a pair of intersecting line segments. One segment is defined as the line from the head of the aircraft to the tail. The other one is the line between left and right wings. Fig. \ref{Figure 3} shows an overview of our framework. 104-\textit{Hourglass}\cite{newell2016stacked} is selected as the backbone network of our model for its powerful capability of feature fusion. Two prediction modules in parallel after the backbone network are in charge of predicting two line segments respectively with the output form of heatmaps. When two aircraft are parked close in an image, their corresponding line segments may be connected together to form a single line segment wrongly, which we call line segment adhesion problem. In order to solve this problem, a third prediction module in parallel is added for the prediction of four endpoints of two target line segments, which acts as a basis for inspecting the phenomenon of adhesion and locating the position of adhesion line segment. Furthermore, in order to provide context of the  direction of aircraft, a fourth prediction module in parallel is designed to output the heatmap of the head keypoint of each aircraft. With these four heatmaps provided, simple post-processing algorithms are applied and finally three forms of bounding boxes can be generated by \textit{X-LineNet} simultaneously. In this paper, we name these four heatmaps as heatmap $A,B,C$ and $D$ in order.

\begin{figure}[!t]
	\centering
	\includegraphics[width=8.5cm,height=6.3cm]{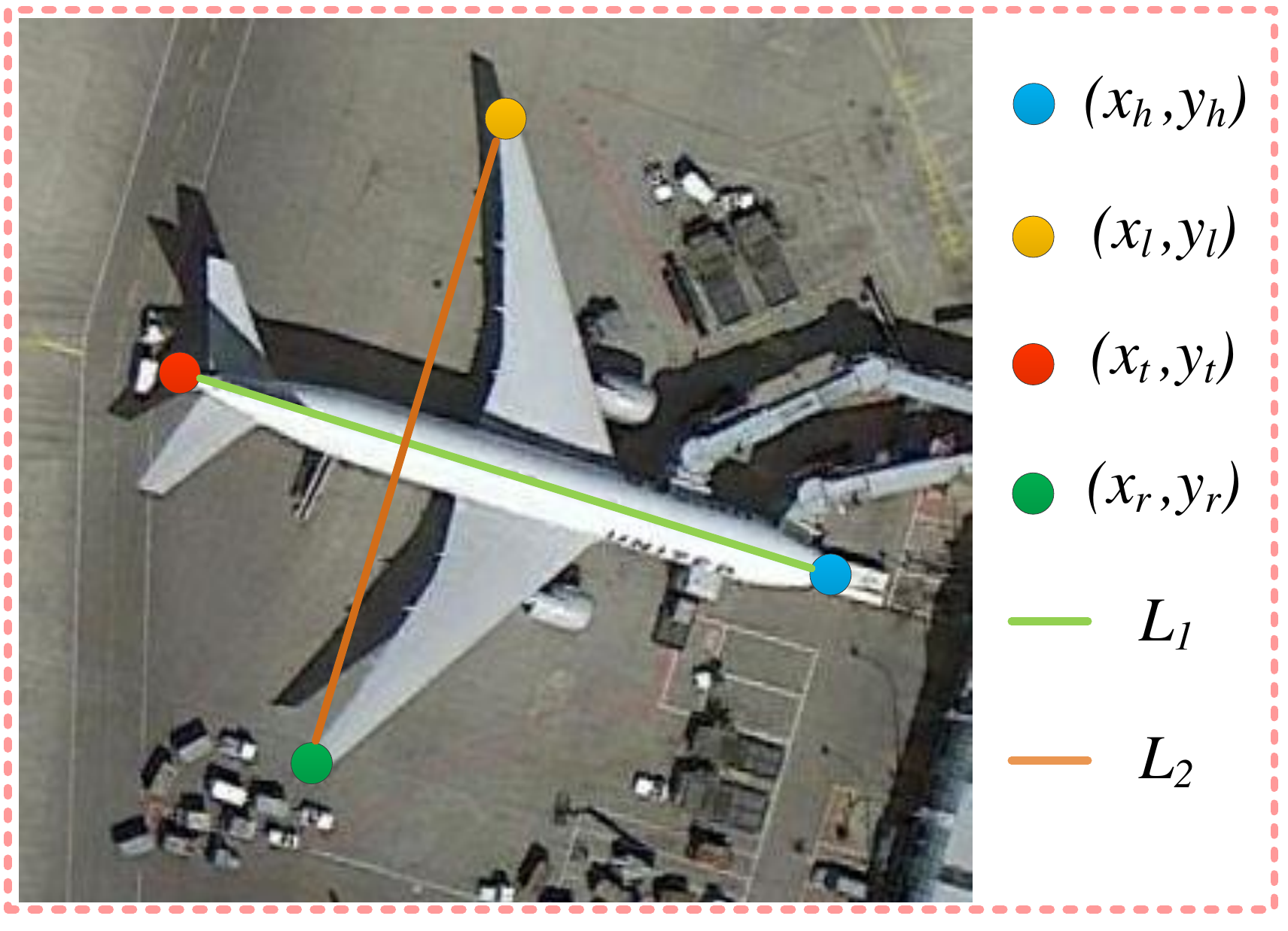}
	\caption{Keypoints and line segments. The four points in the figure correspond to $(x_h,y_h) ,(x_l,y_l), (x_t,y_t), (x_r,y_r)$ respectively. The two segments correspond to $L_1$ and $L_2$ respectively.}
	\label{Figure 4}
\end{figure}

\subsection{Detection of Line Segments and Keypoints} \label{section: 3.2}
Based on  the feature map extracted by the backbone network, four sets of heatmaps will be generated in parallel after a series of layers in prediction modules. Supposing the shape of the original input image is $H\times W$, the size of each heatmap is $\frac{H}{D}\times \frac{W}{D}$ where $D$ corresponds to the output stride. Each pixel value $y\in[0,1]^{\frac{H}{D}\times \frac{W}{D}}$ in heatmaps represents the confidence of being judged as positive sample.

As shown in Fig. \ref{Figure 4}, let  $(x_h,y_h)$ be the head keypoint of the aircraft and $(x_t,y_t)$ be the keypoint of tail. And let $(x_l,y_l)$ and $(x_r,y_r)$ be the keypoints of two wings of aircraft respectively.   
For heatmap $A$ and $B$, We define the segment from the head of each aircraft to the tail part as $L_1[(x_h,y_h),(x_t,y_t)]$ , and $L_2[(x_l,y_l),(x_r,y_r)]$ as the segment between left wings and right wings. All pixels value within $L_1$ and $L_2$ set are fixed to the value of $1$ as positive samples, and other pixels are regarded as the background with their values setting to $0$. In the process of training, objective function is set to the pixel-level logistic regression with a modified \textit{Focal Loss}\cite{lin2017focal} named $\mathcal{L}_{ls}$ which means the loss of line segments:
\vspace{2 ex}
\begin{eqnarray}
\mathcal{L}_{ls} = -\frac{1}{N}\sum_{xy}
\begin{cases}
(1-\acute{{V_{xy}}})^\gamma\log \acute{{V_{xy}}},  & \text{\ if $V_{xy}$ = 1} \\
\acute{{V_{xy}}}^\gamma\log(1-\acute{{V_{xy}}}), & \text{\ if $V_{xy}$ = 0}
\end{cases}
\vspace{2 ex}
\end{eqnarray}
where $N$ is the number of aircraft, $\acute{V_{xy}}$ represents the pixel value at the coordinate $(x,y)$ in heatmap $A$ and $B$, and $V_{xy}$ corresponds to the ground truth. $\gamma$ is a hyper-parameter in original \textit{Focal Loss}, here we set it to 2 in our model.  

The adding of heatmap $C$ is to deal with the line segment adhesion problem ({\color{red}\textit{Section \ref{section: 3.1}}}).  When one aircraft is parked close to another in an image, this problem may occur in any endpoints of two target line. In order to locate the position of adhesion line segment, we let heatmap $C$ output all the four endpoints of each aircraft, so the ground truth of heatmap $C$ is the four endpoints $(x_h,y_h) ,(x_l,y_l), (x_t,y_t), (x_r,y_r)$ of each aircraft with values setting to $1$.

We design the fourth prediction module to get the direction information of aircraft. One head keypoint of each aircraft can provide the direction information enough, so the ground truth of heatmap $D$ is the keypoint of the head $(x_h,y_h)$ of each aircraft with the value also setting to $1$.

The endpoints loss and head keypoints loss corresponding to heatmap $C$ and heatmap $D$ are named as $\mathcal{L}_{ep}$ and $\mathcal{L}_{hp}$ respectively. They follow the same form of $\mathcal{L}_{ls}$. The total loss of our model can be expressed as: 
\begin{eqnarray}
&&\mathcal{L}oss = \mathcal{L}_{ls} + \alpha{\mathcal{L}_{ep}} + \beta{\mathcal{L}_{hp}}
\end{eqnarray}
where $\alpha$ and $\beta$ are the corresponding weights of losses.


\begin{figure}[!t]
	\centering
	\includegraphics[width=8.8cm]{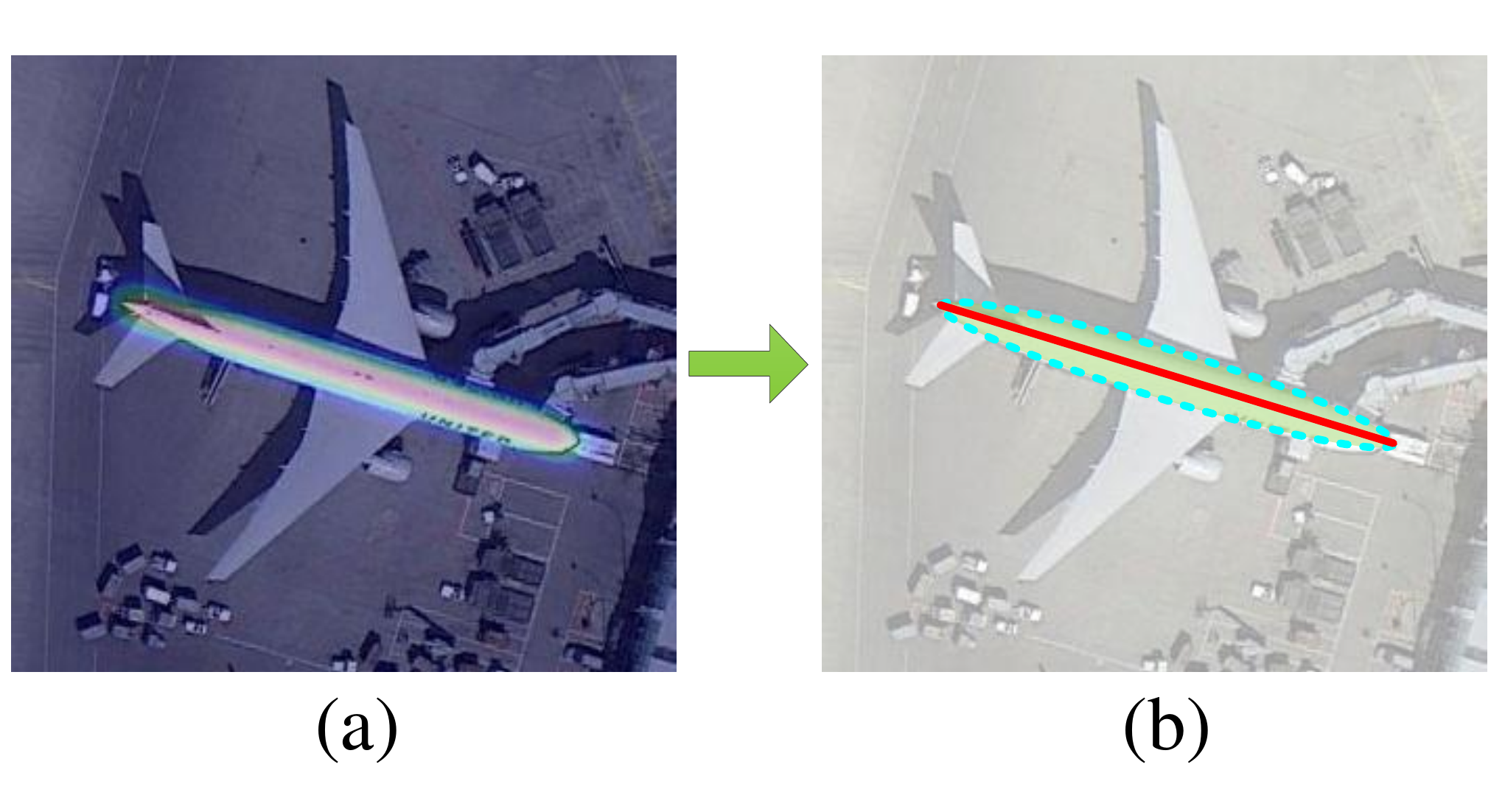}
	\caption{Figure (a) shows the line segments predicted by our model in heatmap is a line area actually. We find the fitting ellipse of each line area and regard the long axis as the line segment we need as shown in Figure (b).}
	\label{Figure 20}
\end{figure}

\subsection{From Heatmaps to Line Segments}\label{section:3.4}

The heatmaps output by our model are the same as confidence maps actually with values $Y\in[0,1]^{H \times W}$. Per pixel near the ground truth in them will be close to $1$ in varying degrees according to the distance, which represents the score of this point considered to be ground truth. Therefore, as shown in Fig. \ref{Figure 20}, that each line segment we need maps a line area in the corresponding heatmap. In the test stage, in order to extract target line segments, a method based on ellipse-fitting algorithm is applied. Concretely, we first use a threshold to transform heatmaps into binary images to filter out the low score pixels. Then we find line areas via the method of extracting connected domains. Last the fitting ellipse can be obtained by fitting each connected domain via ellipse-fitting algorithm. We define the long axis of each ellipse as the desired line segment. 

\begin{figure}[!t]
	\centering
	\includegraphics[width=8.5cm]{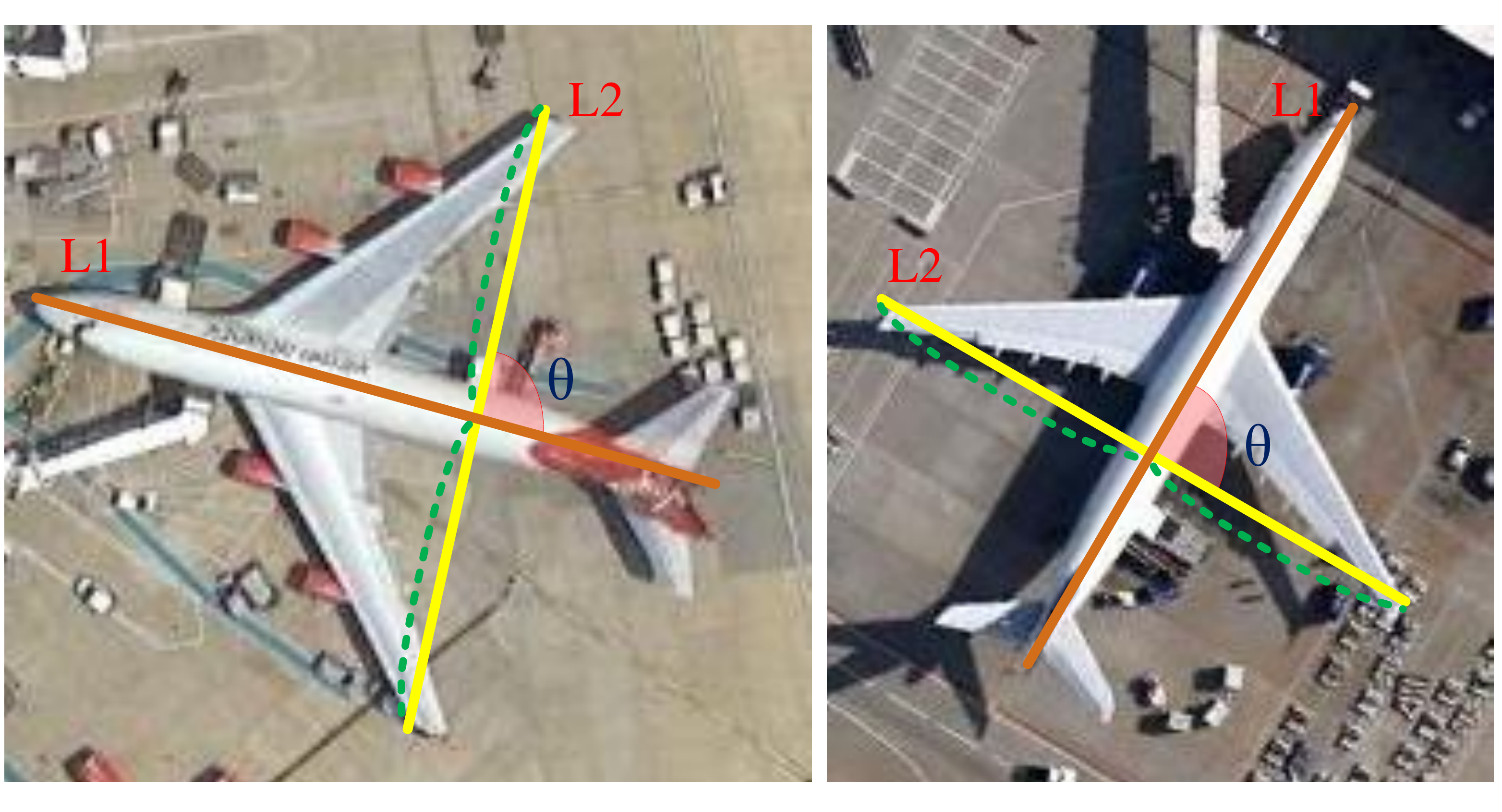}
	\caption{Illustration of the grouping of intersecting line segments.}
	\label{Figure 30}
\end{figure}

\subsection{The Grouping of Line Segments}\label{section:group}

After the line segments extraction, the next step is grouping them which belong to the same aircraft in pairs. We propose two simple clustering rules according to the characteristics of intersecting line segments. Supposing the $L_1$ and $L_2$ represent the line segments extracted in heatmap $A$ and $B$ respectively as shown in Fig. \ref{Figure 30}. The rules are as follows:

(a). $L_2$ is split equally by $L_1$.

(b). The angle $\theta$ between $L_1$ and $L_2$ ranges in certain fields: $\theta \in [60^\circ, 120^\circ]$.

The time complexity of our grouping algorithm is $O(n^2)$, where n is the number of extracted line segments. It is the same as \textit{NMS} which used in anchor-based detectors.

\begin{figure}[!h]
	\centering
	\includegraphics[width=8.5cm]{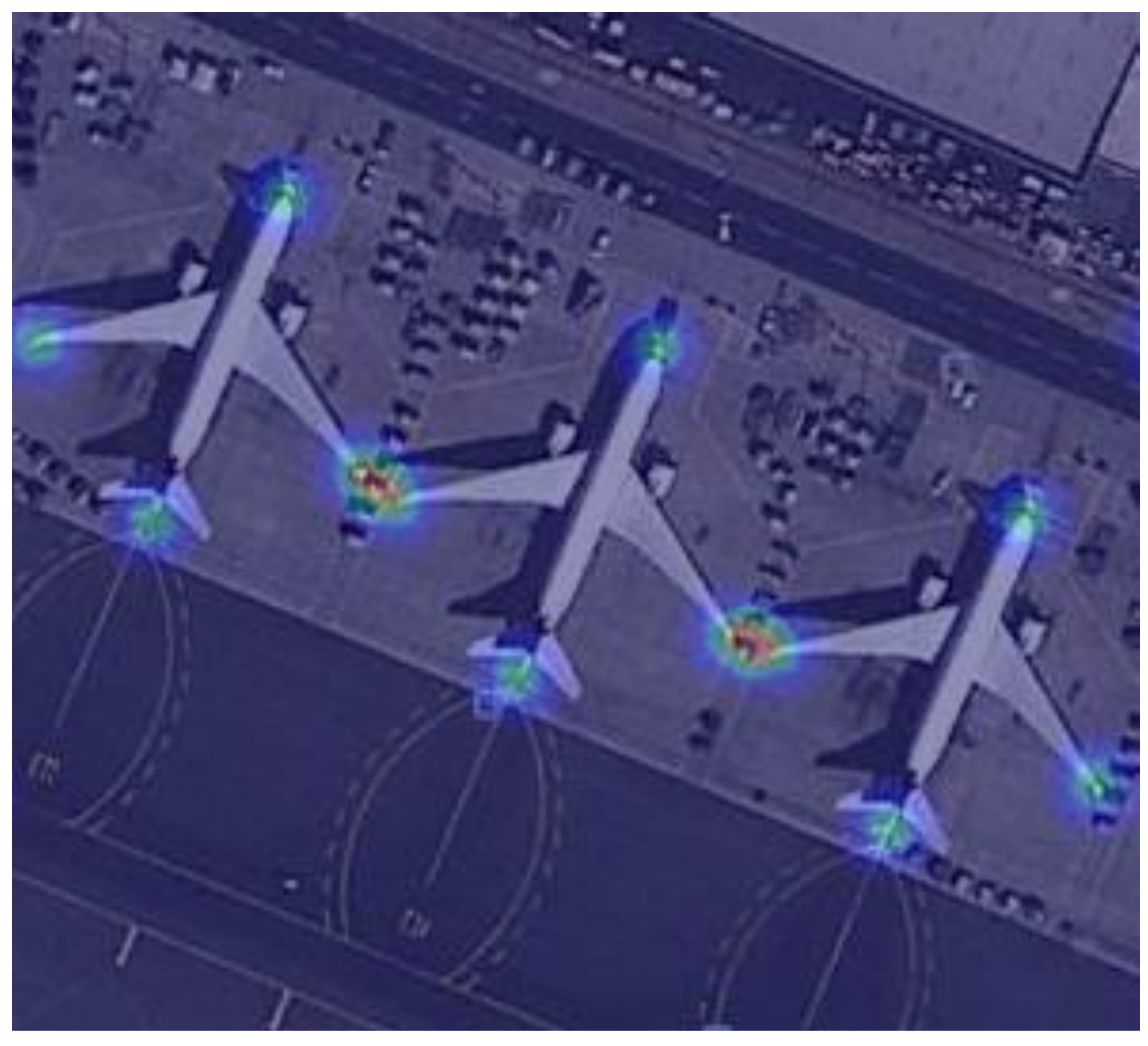}
	\caption{When two aircraft are parked close, the shape and size of corresponding point region of endpoints where the adhesion problem occur are different from other ones in heatmap $C$.}
	\label{Figure 5}
\end{figure}
\subsection{Cutting-line Algorithm for Line Segments Adhesion Problem}\label{section:3.3}
Because the existence of output stride in our model, if two aircraft are close to each other, they may be stick together due to the down sampling. It may lead the adhesion of our target line segments, which we call line segment adhesion problem. We design an algorithm named Cutting-line Algorithm to address this problem. 

The aim of Cutting-line Algorithm is to inspect close aircraft and locates the position of adhesion, then separates the target adhesion line segments. In order to achieve this aim, we introduce heatmap $C$ which outputs four endpoints of each aircraft. The values of pixels in heatmap $C$ will be in the range of $0$ to $1$ and each endpoint is represented by a small point region actually as mentioned in ({\color{red}\textit{Section \ref{section:3.4}}}). As shown in Fig. \ref{Figure 5}, the shape and size of point region in the position of adhesion are different from other ones, which is the key of our Cutting-line Algorithm to distinguish adhesion from non adhesion. We extract each point region via the method of finding connected domain aforementioned, and then introduce the minimum enclosing rectangle of per connected domain, and the shape as well as the size of point region can be represented by the relationship among weight and height of corresponding minimum enclosing rectangle.

Supposing that $\hat{H}$ and $\hat{W}$ correspond to the height and width of the minimum enclosing rectangle, we introduce two relaxation conditions for the situation of non adhesion:

\begin{eqnarray}
&\hat{H}\times \hat{W} < \alpha & \text{\ \ \  $\alpha$ is threshold 1}\\
&\max(\hat{H}, \hat{W}) \div \min(\hat{H}, \hat{W}) < \beta & \text{\ \ \ $\beta$ is threshold 2}
\end{eqnarray}

\begin{figure}[!h]
	\centering
	\includegraphics[width=8.8cm]{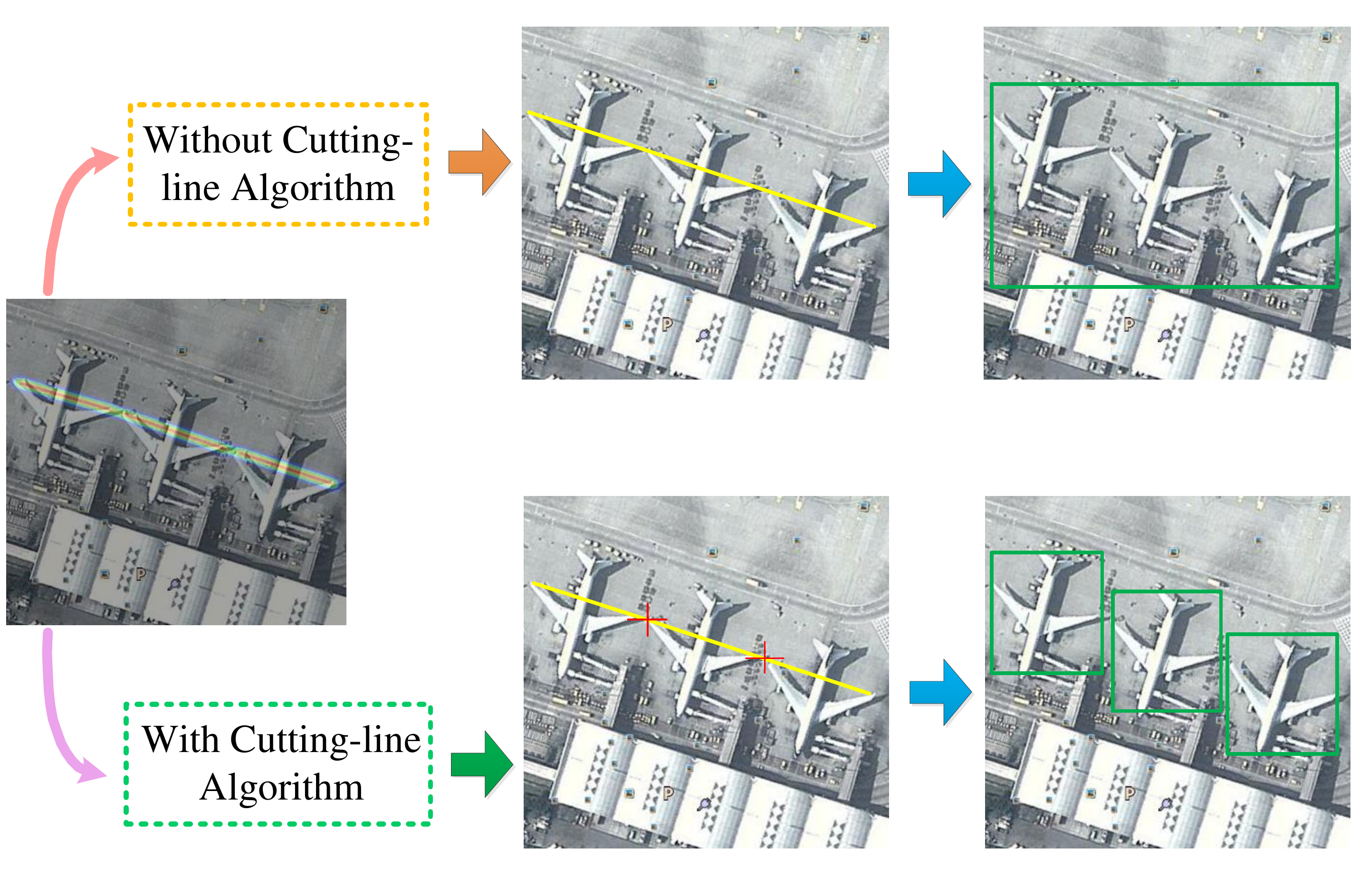}
	\caption{Illustration of the effect of Cutting-line Algorithm.}
	\label{Figure 6}
\end{figure}

Then, we can find out the minimum enclosing rectangle of connected domain that do not meet the relaxation conditions above, and there is a great probability the adhesion problem occur.
With the adhesion connected domains be screened, we can separate the adhesion line segments in heatmap $A$ and $B$ according to the positions of these adhesions in heatmap $C$. Algorithm \ref{Algorithm 1} shows this specific process, and Fig. \ref{Figure 6} shows the effect of Cutting-line Algorithm.

\begin{algorithm}[!h] 
	
	\caption{ Cutting-line Algorithm.}  
	\label{Algorithm 1}
	\begin{algorithmic}[1]  
		\State let $D$ be a collection of 4-connected domains  in heatmap $C$
		\For{each $d\in D$}  
		\State obtain the minimum external rectangle $R$ of $d$;
		\State let $\hat{H}$ and $\hat{W}$ be the length and width of $R$;
		\State let $p_c$ be the center point of $R$, and $P$ is an empty array;
		\State let $\alpha$ and $\beta$ be two constant thresholds.
		\If {$\hat{H}\times \hat{W} < \alpha$ \textbf{or} $\max(\hat{H}, \hat{W}) \div \min(\hat{H}, \hat{W}) < \beta$}
		\State Continue
		\Else
		\State $P \gets p_c$
		\EndIf
		\EndFor 
		\For{each $p\in P$ in heatmap $A$ and $B$}
		\State  per pixel of $L_a[(p_x-k/2,p_y),(p_x+k/2,p_y)] = 0$
		\State  per pixel of $L_b[(p_x,p_y-k/2),(p_x,p_y+k/2)] = 0$
		\EndFor
		
	\end{algorithmic}  
\end{algorithm}

\begin{figure*}[t]
	\centering
	\includegraphics[width=18.5cm]{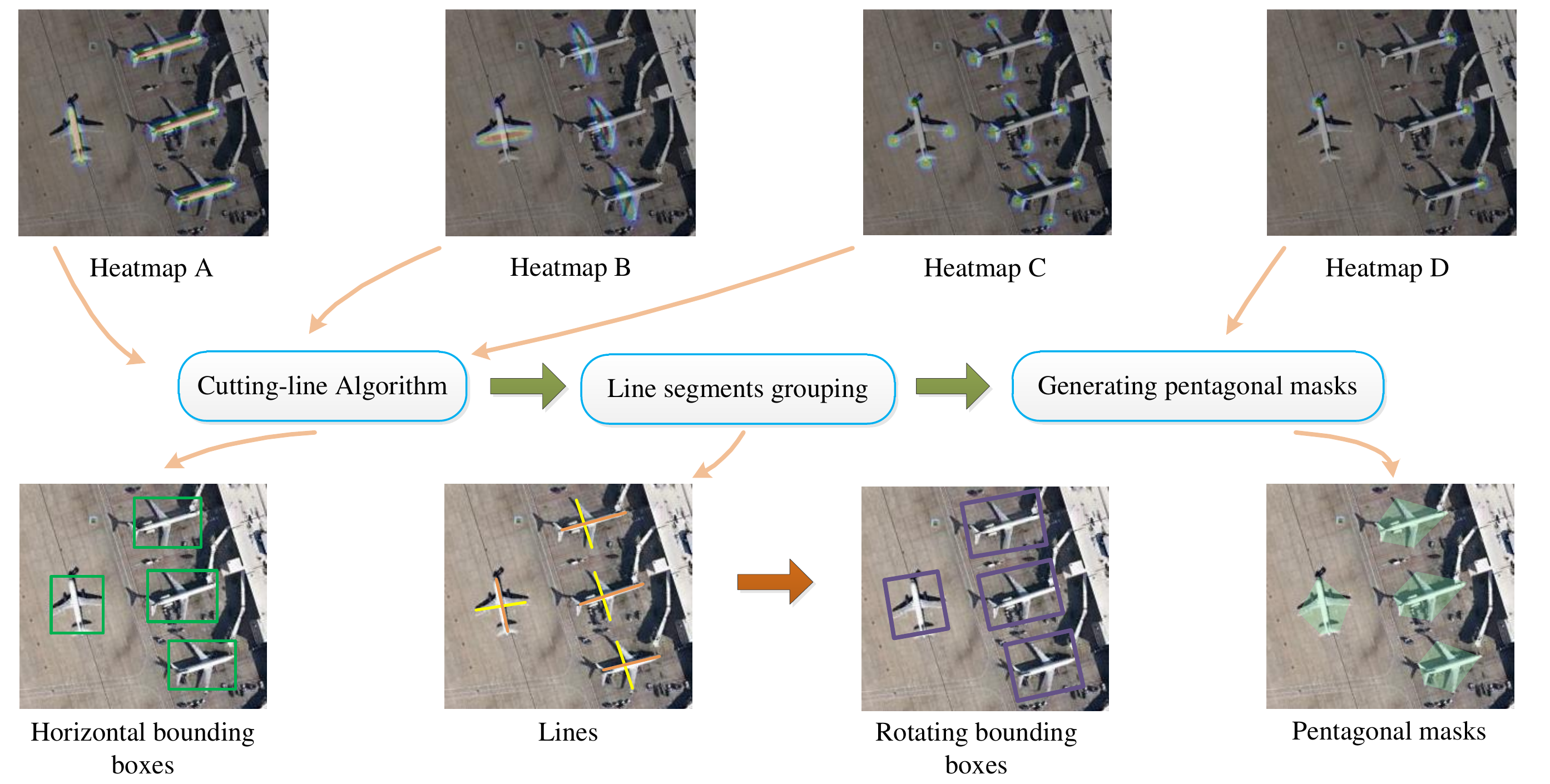}
	\caption{The process of generating bounding boxes. }
	\label{Figure 7}
\end{figure*}

\subsection{The Generation of Pentagonal Mask}
\textit{X-LineNet} can generate three forms of bounding boxes as shown in Fig. \ref{Figure 7}. For the purpose of generating the pentagonal mask with extra direction context of aircraft, we need to identify the specific meaning of these four endpoints of two target line segments after the clustering in {\color{red}\textit{Section \ref{section:group}}}. The output of heatmap $D$ is to achieve this aim. Heatmap $D$ provides the position of head point of each aircraft, and we can match it via \textit{Euclidean distance} to two endpoints of $L_1$ which defined as the line segment between the head of aircraft to the tail. The endpoint of $L_1$ with the largest matching value is regarded as the head point and the other one is the tail. After the head and tail point are found by us, other two keyponits of $L_2$ that between two wings can be gained easily. It is noteworthy that the bottom of the pentagonal mask is a line, so we define the line as the direction parallel to $L_2$ and the length of it are fixed as $\frac{1}{5}$ length of  $L_2$. With the above operation, the pentagonal mask can be obtained.

\begin{table*}[t]\normalsize
	\centering
	\renewcommand{\captionlabelfont}{\bfseries}
	\setlength{\tabcolsep}{2.5mm}{
		\caption{State-of-the-art comparisons.  The input size is the resolution fed into the model during training. S means single scale and M means multi-scale. The RRPN is the \textit{Rotation Region Proposal Networks}. The R2CNN indicates \textit{Rotational Region CNN}. The R-DFPN means the \textit{Rotation Dense Feature Pyramid Networks}.}
		\label{table 1}
		\begin{tabular}{ccccccccc}
			\toprule  
			\toprule  
			\textbf{Models}&\textbf{Backbone}&\textbf{Input\ size}&$AP^1$&$AP^2$&$AP^m$&$AP^1$&$AP^2$&$AP^m$ \\
			
			\midrule
			\midrule  
			\rowcolor{gray!20}
			\multicolumn{1}{c}{\textbf{Horizontal Bounding Box}}&\multicolumn{2}{c}{}&\multicolumn{3}{c}{\textbf{Training on \textit{Aircraft-KP}}}&\multicolumn{3}{c}{\textbf{Training on \textit{ UCAS-AOD}}}\\
			\midrule
			\textbf{Two-stage detectors}\\
			Faster R-CNN\cite{ren2015faster}&vgg16&1200$\times$600&85.9\%&85.6\%&85.7\%&86.4\%&85.2\%&85.8\%\\
			Faster R-CNN\cite{he2016deep}&ResNet-50&1200$\times$600&88.7\%&88.3\%&88.5\% &88.4\%&88.2\%&88.3\%\\
			Faster R-CNN\cite{he2016deep}&ResNet-101&1200$\times$600&90.1\%&90.5\%&90.3\%&89.8\%&90.4\% &90.1\%\\
			Faster R-CNN+FPN\cite{lin2017feature}&ResNet-101&1000$\times$600&91.3\%&91.2\%&91.2\%&92.4\%&91.5\%&91.9\%\\
			Cascade R-CNN\cite{cai2018cascade}&ResNet-101&1024$\times$512&\textbf{92.8\%}&\textbf{92.1\%}&\textbf{92.4\%}&\textbf{93.1\%}&\textbf{91.8}\%&\textbf{92.4\%}\\
			
			\midrule  
			\textbf{One-stage detectors}\\
			YOLO9000\cite{redmon2017yolo9000}&DarkNet-19&544$\times$544&82.2\%&81.9\%&82.1\%&82.4\%&81.4\%&81.9\%\\
			YOLOv3\cite{redmon2018yolov3}&DarkNet-53&608$\times$608 &86.4\%&86.2\%&86.3\%&85.4\%&85.5\%&85.4\%\\
			SSD\cite{liu2016ssd}&ResNet-101&513$\times$513&89.6\%&89.4\%&89.5\%&88.7\%&89.4\%&89.0\%\\
			RetinaNet\cite{lin2017focal}&ResNet-101&512$\times$512&90.1\%&90.3\%&90.2\%&89.8\%&90.9\% &90.3\%\\
			ConerNet(S)\cite{law2018cornernet}&104-Hourglass&511$\times$511&76.5\%&75.4\%&75.9\%&76.2\%&77.4\% &76.8\%\\
			ConerNet(M)\cite{law2018cornernet}&104-Hourglass&511$\times$511&76.7\%& 77.3\%&77.0\%&76.8\%&77.4\%&77.1\%\\
			ExtremeNet\cite{zhou2019bottom}&104-Hourglass&511$\times$511&78.3\%&77.6\% &77.9\%&78.3\%&77.2\%&77.7\% \\
			Our&104-Hourglass&511$\times$511 &\textbf{92.5}\%&\textbf{92.1}\%&\textbf{92.3}\%&\textbf{91.9\%} &\textbf{91.2}\%&\textbf{91.5}\%\\
			
			\midrule  
			\midrule  
			\rowcolor{gray!20}
			\multicolumn{1}{c}{\textbf{Rotating Bounding Box}}&\multicolumn{2}{c}{}&\multicolumn{3}{c}{\textbf{Training on \textit{Aircraft-KP}}}&\multicolumn{3}{c}{\textbf{Training on \textit{UCAS-AOD}}}\\
			\midrule
			RRPN\cite{ma2018arbitrary}&ResNet-50&1000$\times$600&90.1\%&- &- &90.4\%&- &-\\
			R2CNN\cite{jiang2017r2cnn}&ResNet-101&1000$\times$600&90.7\%&- &- &90.9\%&- &-\\
			R-DFPN\cite{yang2018automatic}&ResNet-101&1000$\times$600&91.0\%&- &- &90.6\%&- &-\\
			Our&104-Hourglass&511$\times$511&\textbf{92.8\%} &-&- &\textbf{91.3\%}&-&-\\
			
			\bottomrule  
	\end{tabular}}
	
\end{table*}

\section{Experiments}\label{section:4}

\subsection{Datatsets}\label{section:4.1}
\begin{figure}[!h]
	\centering
	\includegraphics[width=9cm]{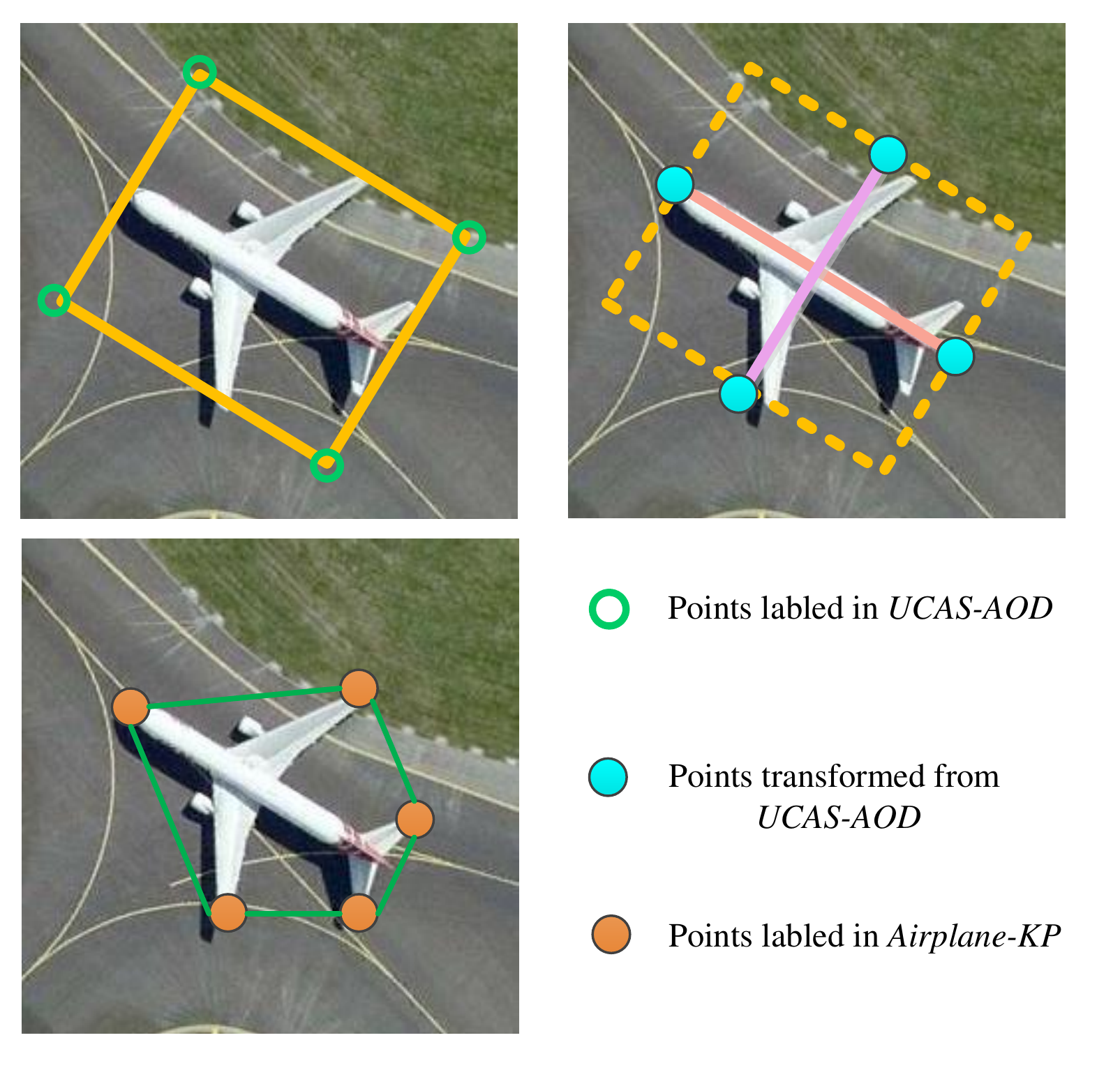}
	\caption{The label formats of \textit{UCAS-AOD} and \textit{Aircraft-KP} dataset. }
	\label{Figure 8}
\end{figure}

There are two datasets in our experiments. The first one is \textit{UCAS-AOD}\cite{zhu2015orientation}, which has high quality remote sensing images and sufficient rotating bounding box annotations of aircraft. The format of annotations in \textit{UCAS-AOD} is shown in Fig. \ref{Figure 8}. Considering there are no direct line segment annotations of aircraft in current open dataset, the line segments that our \textit{X-LineNet} need have to be transformed from the rotating bounding box annotations.  We take the median line of each rotating rectangle box as the two line segments that need to be predicted by our model. However, we find the directly converted annotation is not accurate enough, such as the line segments between left and right wings is not always located at the middle line of the rotating bounding box, which may be bad for our model. In order to explore the full capability of \textit{X-LineNet}, we release a new aircraft keypoint dataset named \textit{Aircraft-KP}. Images of training sets and testing sets in \textit{Aircraft-KP} are the same as \textit{UCAS-AOD} and we just provide a whole new set of labels. In \textit{Aircraft-KP}, each aircraft is labeled as five keypoints along the anti-clockwise direction by us for better versatility. Concretely, these five endpoints are defined as following order: head of aircraft, left wing, left tail, right tail, right wing as shown in Fig. \ref{Figure 8}. Note that we define the midpoint of both sides of the tail part as the tail point in our experiments. \textit{Aircraft-KP} is available at \ \url{https://github.com/Ucas-HaoranWei}.

\subsection{Training and Testing Details}\label{section:4.2}
There are $7482$ targets aircraft in $1000$ images totally in \textit{UCAS-AOD} and \textit{Aircraft-KP}. We randomly select $800$ images used for training, leaving other $200$ for testing. In order to test the generalization performance of \textit{X-LineNet}, \textit{NWPU VHR-10}\cite{cheng2016learning} dataset is merged to our test experiment. It contains about $700$ aircraft in $80$ images. Our experiments are performed on one \textit{RTX-2080Ti GPU}, \textit{I9-9900K CPU }with \textit{PyTorch 1.0}\cite{paszke2017automatic}.

We train \textit{X-LineNet} on the new datasets(\textit{Aircraft-KP}) and the original datasets(\textit{UCAS-AOD}) respectively. During the training phase, we set the input resolution to $511 \times 511$ and the output resolution to $128 \times 128$ following settings in \textit{CornerNet}\cite{law2018cornernet}. In order to get rid of the risk of overfitting, some data argumentation methods are applied, including random horizontal flipping, vertical flipping and color dithering. Adam\cite{kingma2014adam} is selected as the optimizer for \textit{X-LineNet} with  learning rate of $0.0025$. We train our network from scratch for $40000$ iterations with the batch size setting to $4$. The total loss({\color{red}\textit{Section \ref{section: 3.2}}}) of \textit{X-LineNet} is as follows : 
\begin{eqnarray}
&&\mathcal{L}oss = \mathcal{L}_{ls} + \alpha{\mathcal{L}_{ep}} + \beta{\mathcal{L}_{kp}}
\end{eqnarray}
where both $\alpha$ and $\beta$ are set to 0.5 in training process.

During the stage of test, We keep the original resolution of image instead of resizing it to a fixed size and all of our test results are obtained from a single scale. In our experiments, we test the output form of horizontal and rotating bounding box respectively. We use the average precision(\textit{AP}) as the evaluation metrics which are defined in \textit{PASCAL VOC Challenge}\cite{everingham2010pascal}. We define $AP^1$ as the test result on \textit{UCAS-AOD} and \textit{Aircraft-KP}, $AP^2$ as the result on \textit{NWPU VHR-10 } and their average as $AP^m$. We choose the default parameters in \textit{PASCAL VOC} with \textit{IoU}(Intersection over Union) which is $0.5$ during testing. Note that \textit{NWPU VHR-10 dataset} is only added to the horizontal bounding box test process because it lacks the annotations of the rotating bounding box. The thresholds are used to transform heatmaps to binary images are all setting to $0.3$. In Cutting-line Algorithm ({\color{red}\textit{Section \ref{section:3.3}}}), the $\alpha$ and $\beta$ of relaxation conditions are setting to $100$ and $1.5$ respectively. And number of cutting pixel $k$ is setting to $10$ in our experiments.

\begin{figure*}[!t]
	\centering
	\includegraphics[width=18cm]{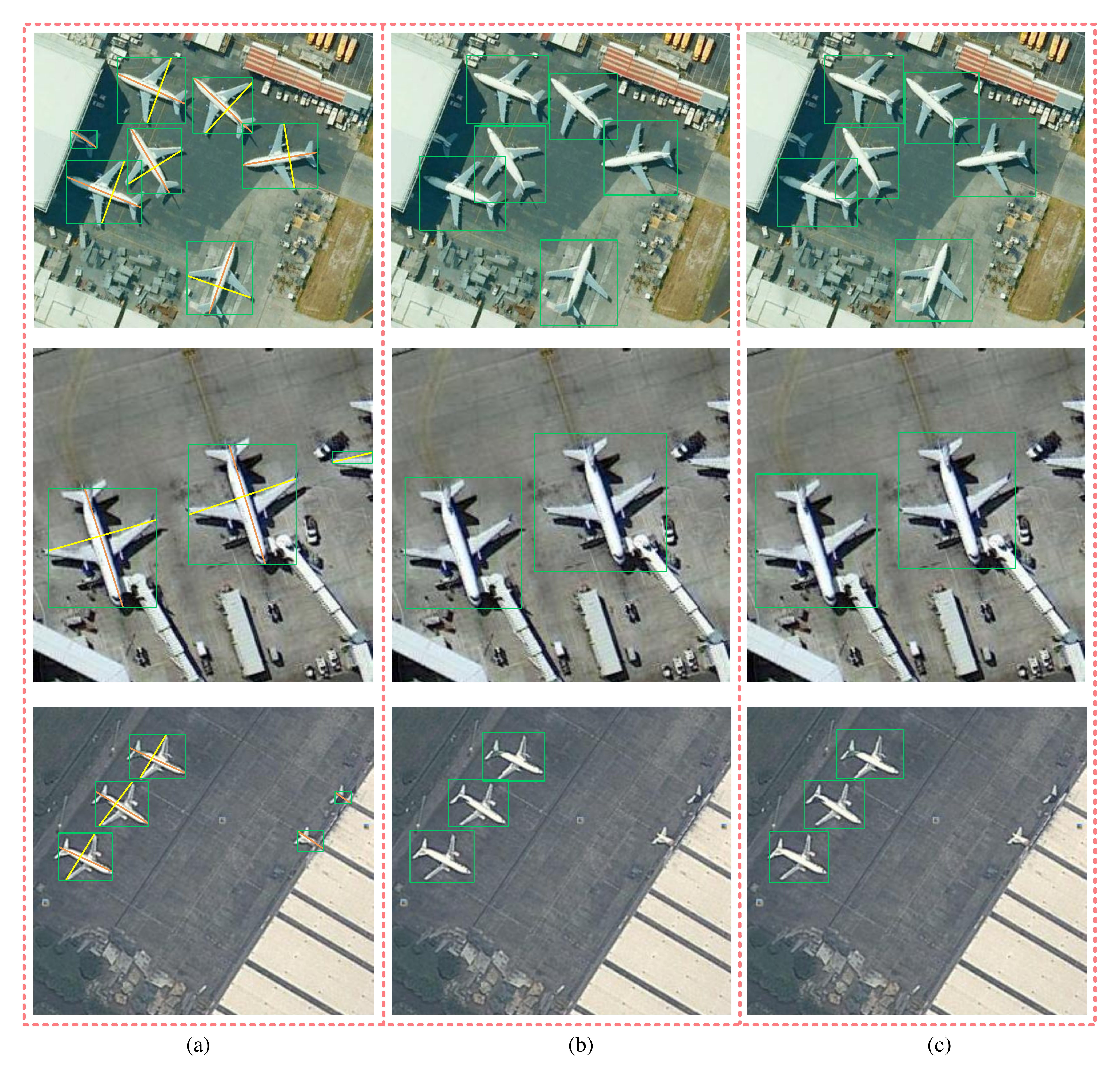}
	\caption{The aircraft can still be detected by our model in an occasion that it is severely shielded. Figure (a) is the effect of \textit{X-LineNet}. Figure (b) and (c) are effects of \textit{Faster R-CNN+FPN} and \textit{Cascade R-CNN} respectively. }
	\label{Figure 9}
\end{figure*}

\begin{figure*}[!t]
	\centering
	\includegraphics[width=18.2cm]{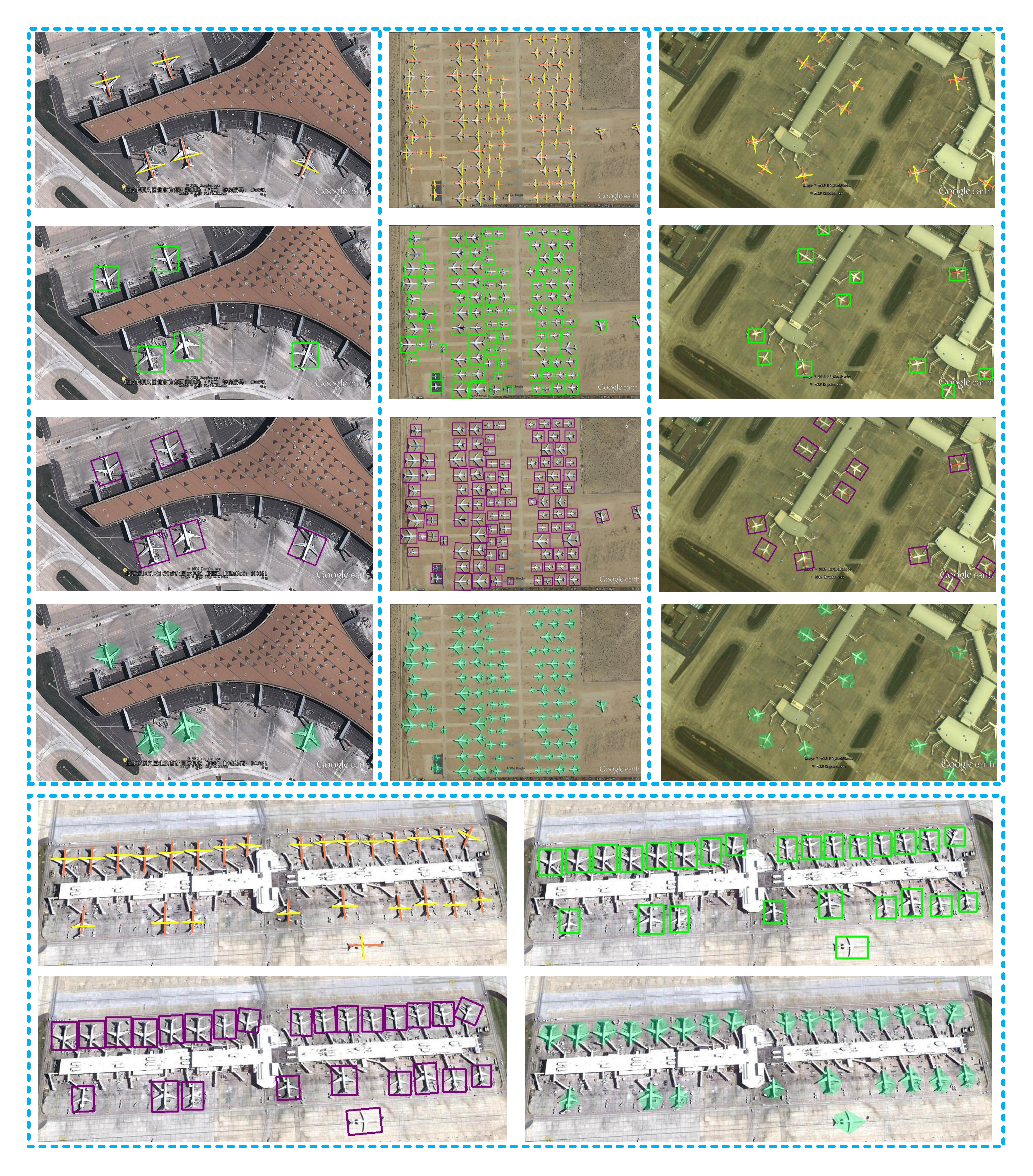}
	\caption{Qualitative results output by \textit{X-LineNet}. }
	\label{Figure 10}
\end{figure*}

\subsection{Comparisons with State-of-the-art Frameworks}
In this part, we first prove the advancement of \textit{X-LineNet} on the new dataset \textit{Aircraft-KP}. For the sake of testifying the satisfactory performance of our model does not depend on the introduction of more precise annotations in \textit{Aircraft-KP}, we then carry on the experiment on original \textit{UCAS-AOD}.

\subsubsection{Experiment on \textit{Aircraft-KP}}\label{section:4.3.1}
We compare the performance of \textit{X-LineNet} with other state-of-the-art detectors in aircraft detection. Annotations of training sets used by other detectors are converted from \textit{Aircraft-KP}. Table \ref{table 1} shows the results. Our model achieves an $AP^m$ of 92.3\%, outperforming all reported one-stage object detectors and going far beyond other bottom-up models. \textit{X-LineNet} is even competitive to the most advanced two-stage detectors. 

For the output form of rotating bounding box, we select several state-of-the-art  models used for detecting oriented objects in remote sensing images for comparison. As shown in Table \ref{table 1}, our \textit{X-LineNet} improves about $2\%$ $AP^1$ compared with other top-down detectors which use a design of rotated anchors.

It is noteworthy that  the aircraft can still be detected by our model in an occasion that it is severely shielded or not fully displayed at the edge of the image due to the fact that our network enhances the learning of target's visual grammar information. The effects are shown in Fig. \ref{Figure 9}.

\subsubsection{Experiment on \textit{UCAS-AOD}}\label{section:4.3.2}
In order to prove the excellent performance of \textit{X-LineNet} is not relied on the additional annotations in \textit{Aircraft-KP}, we conduct another experiment on the original \textit{UCAS-AOD}. As known in {\color{red}\textit{Section \ref{section:4.1}}},
\textit{X-LineNet} can also be trained with the data converted from the orginal annotations of \textit{UCAS-AOD} directly instead of \textit{Aircraft-KP} labeled by us. We define the two median lines of the original labeled rotating bounding box as the segments that need to be detected as shown in Fig. \ref{Figure 8}. It is worth mentioning that our model can still achieve good accuracy with the converted training sets without additional manual tagging. State-of-the-art detectors used for comparison are the same as {\color{red}\textit{Section \ref{section:4.3.1}}}. It's just that they are trained directly through the original annotations of \textit{UCAS-AOD}.  Table \ref{table 1} also shows the contrast results. \textit{Aircraft-KP} gives an $AP^m$ improvement of $0.8\%$ than the training data converted from \textit{UCAS-AOD} directly. Because the endpoints of segments in \textit{Aircraft-KP} are explicit and clearer which help the Cutting-line Algorithm ({\color{red}\textit{Section \ref{section:3.3}}}) be implemented better. The validation experiment we conducted is shown in {\color{red}\textit{Section \ref{section:4.4.2}}}.

\subsection{Ablation Studies}

\subsubsection{104-Hourglass with Anchors}
Unlike \textit{X-LineNet}, current top-down and anchor-based object detectors  do not use \textit{104-Hourglass}  as a backbone. In order to show the excellent performance of \textit{X-LineNet} is not only dependent on the contribution of \textit{104-Hourglass}, we replace the backbone of an anchor-based and one-stage detector \textit{RetinaNet}\cite{lin2017focal} with \textit{104-Hourglass}. The setting of anchors is in multi-scale in unsampling stage of \textit{104-Hourglass}. The evaluation metrics are \textit{AP} of horizontal bounding boxes. Table \ref{table 2} shows that with \textit{104-Hourglass} as the backbone network, our \textit{X-LineNet} improves $AP^m$ by 9.9\% than the anchor-based model. In conclusion, compared with anchor-based methods, the performance improvement of \textit{X-LineNet} is not due to the different backbone networks.

\begin{table}[!h]\normalsize
	\centering	
	\renewcommand{\captionlabelfont}{\bfseries}
	\setlength{\tabcolsep}{3.2
		mm}{
		\caption{Comparisons of \textit{X-LineNet} and \textit{104-Hourglass} with anchors.}
		\label{table 2}
		\begin{tabular}{cccc}
			\toprule  
			\rowcolor{gray!20}
			\textbf{Methods}
			&$AP^1$&$AP^2$&$AP^m$ \\ 
			\midrule  
			X-LineNet&\textbf{92.5}\%&\textbf{92.1}\%&\textbf{92.3}\% \\
			104-Hourglass+anchors&82.2\%&82.6\%&82.4\% \\
			\bottomrule  
	\end{tabular}}
	
\end{table}

\subsubsection{Different Backbone Networks}

To verify the necessity of \textit{104-Hourglass} network for our model, we replace the \textit{104-Hourglass} with \textit{ResNet-101+FPN}\cite{lin2017feature} which is more commonly used in popular detectors at present. We only use the final output of \textit{FPN} for predictions. The model is trained following the same training procedure in {\color{red}\textit{Section \ref{section:4.2}}}. The detection metrics are also \textit{AP} of horizontal bounding boxes. Table \ref{table 4} shows the contrast results.
The experimental results show that \textit{X-LineNet} with \textit{104-Hourglass} outperforms the \textit{X-LineNet} with \textit{ResNet-101+FPN} by 9.4\% $AP^m$. Therefore, the matching of our model with \textit{104-Hourglass} network is very necessary.

\begin{table}[!h]\normalsize
	\centering	
	\renewcommand{\captionlabelfont}{\bfseries}
	\setlength{\tabcolsep}{3.5mm}{
		\caption{Comparisons of Different Backbone Networks.}
		\label{table 4}
		\begin{tabular}{cccc}
			\toprule  
			\rowcolor{gray!20}
			\textbf{Different Backbones}&$AP^1$&$AP^2$&$AP^m$ \\ 
			\midrule  
			104-Hourglass&\textbf{92.5}\%&\textbf{92.1}\%&\textbf{92.3}\% \\
			ResNet-101+FPN&83.2\%&82.7\%&82.9\% \\
			\bottomrule  
	\end{tabular}}
	
\end{table}

\begin{table}[!h]\normalsize
	\centering	
	\renewcommand{\captionlabelfont}{\bfseries}
	\setlength{\tabcolsep}{3mm}{
		\caption{Effects of Cutting-line Algorithm.}
		\label{table 3}
		\begin{tabular}{cccc}
			\toprule  
			\midrule
			\multicolumn{4}{c}{\textbf{With or without Cutting-line Algorithm}}\\  
			\midrule  
			\midrule
			\rowcolor{gray!20}
			\textit{Training on Aircraft-KP}&$AP^1$&$AP^2$&$AP^m$ \\
			\midrule  
			without cutting&91.2\%&90.7\%&90.9\% \\
			with cutting&\textbf{92.5}\%&\textbf{92.1}\%&\textbf{92.3}\% \\
			\midrule  
			\rowcolor{gray!20}
			\textit{Training on UCAS-AOD}&$AP^1$&$AP^2$&$AP^m$\\
			\midrule  
			without cutting&91.0\%&90.5\%&90.7\% \\
			with cutting&\textbf{91.9}\%&\textbf{91.2}\%&\textbf{91.5}\% \\
			\bottomrule  
	\end{tabular}}
	
\end{table}

\subsubsection{Effects of Cutting-line Algorithm}\label{section:4.4.2}

In order to verify the significance of Cutting-line Algorithm ({\color{red}\textit{Section \ref{section:3.3}}}), we compare the performance of our method with Cutting-line Algorithm and without it respectively, on both \textit{UCAS-AOD} and \textit{Aircraft-KP} datasets. Evaluation metrics are \textit{AP} of horizontal bounding boxes. The experimental results on \textit{Aircraft-KP} and \textit{UCAS-AOD} show that Cutting-line Algorithm gives decent $AP^m$ improvements of $1.4\%$ and $0.8\%$ respectively. It effectively solves the issue of adjacent segments which may occur in an occasian that aircraft are densely placed in a single image. As a result, the accuracy is effectively improved without introducing much cost of computation. Table \ref{table 3} shows the results.  

It is notable that \textit{Aircraft-KP} with Cutting-line Algorithm gives an $AP^m$ improvement of $0.8\%$ than the training data converted from \textit{UCAS-AOD} directly because the features of endpoints of the segments in \textit{Aircraft-KP} are more dominant as shown in Fig. \ref{Figure 8}, so that the location of endpoints can be detected more easily by \textit{X-LineNet} ({\color{red}\textit{Section \ref{section:3.3}}}).

\section{Conclusion}\label{section:5}
In conclusion, we present a novel one-stage and anchor-free model named \textit{X-LineNet}, which detects aircraft in remote sensing images based on a bottom-up method. We switch the task of aircraft detection into detecting and grouping a pair of intersecting line segments, which enables our \textit{X-LineNet} to learn richer and specific visual grammar information of aircraft. As a result, our model outperforms state-of-the-art one-stage detectors and is still competitive compared with advanced two-stage detectors in the field of aircraft detection.

\bibliographystyle{IEEEtran} 
\bibliography{main}

%








\end{document}